\documentclass[twoside,11pt]{article}

\usepackage{blindtext}
\usepackage[preprint]{jmlr2e}
\usepackage{amsmath,amsfonts,bm}
\usepackage{dsfont}
\usepackage{algorithm}
\usepackage{algorithmicx}
\usepackage{algpseudocode}
\usepackage{xcolor}  
\def\eqref#1{equation~\ref{#1}}
\def\1{\bm{1}}

\def\vmu{{\bm{\mu}}}
\def\muX{{\vmu_{\scriptscriptstyle \mX}}}

\def\vl{{\bm{l}}}

\def\vx{{\bm{x}}}
\def\vy{{\bm{y}}}
\def\vz{{\bm{z}}}

\def\mA{{\bm{A}}}

\def\mC{{\bm{C}}}

\def\mI{{\bm{I}}}

\def\mK{{\bm{K}}}

\def\mS{{\bm{S}}}
\def\mT{{\bm{T}}}

\def\mW{{\bm{W}}}
\def\mX{{\bm{X}}}
\def\mY{{\bm{Y}}}
\def\mZ{{\bm{Z}}}

\def\mPhi{{\bm{\Phi}}}

\def\mPhiXhat{\hat{\mPhi}_{\scriptscriptstyle \mX}}
\def\mPhiYhat{\hat{\mPhi}_{\scriptscriptstyle \mY}}

\def\mSigma{{\bm{\Sigma}}}

\DeclareMathAlphabet{\mathsfit}{\encodingdefault}{\sfdefault}{m}{sl}
\SetMathAlphabet{\mathsfit}{bold}{\encodingdefault}{\sfdefault}{bx}{n}

\newcommand{\E}{\mathbb{E}}

\newcommand{\normltwo}{L^2}

\DeclareMathOperator{\Tr}{Tr}

\algnewcommand\algorithmicinput{\textbf{Input:}}
\algnewcommand\Input{\item[\algorithmicinput]}
\algnewcommand\algorithmicoutput{\textbf{Output:}}
\algnewcommand\Output{\item[\algorithmicoutput]}

\newcommand{\Real}{\mathbb{R}}

\newcommand{\Kx}{\mK_{\scriptscriptstyle \mX}}
\newcommand{\Ky}{\mK_{\scriptscriptstyle \mY}}

\newcommand{\Kz}{\mK_{\scriptscriptstyle\mZ}}

\newcommand{\CP}{C_{\scriptscriptstyle P}}
\newcommand{\CQ}{C_{\scriptscriptstyle Q}}
\newcommand{\CM}{C_{\scriptscriptstyle M}}
\newcommand{\Cx}{\mC_{\scriptscriptstyle \mX}}

\newcommand{\hatCx}{\hat{\mC}_{\scriptscriptstyle \mX}}
\newcommand{\hatCy}{\hat{\mC}_{\scriptscriptstyle \mY}}
\newcommand{\Cy}{\mC_{\scriptscriptstyle \mY}}
\newcommand{\Cz}{\mC_{\scriptscriptstyle \mZ}}

\renewcommand{\d}[1]{\ensuremath{\operatorname{d}\!{#1}}}
\usepackage{mathtools}

\usepackage[normalem]{ulem}
\newcommand\redsout{\bgroup\markoverwith{\textcolor{red}{\rule[0.5ex]{2pt}{0.4pt}}}\ULon}

\usepackage{booktabs, hhline, wrapfig} 
\usepackage{subfigure}
\usepackage{lastpage}
\jmlrheading{25}{2024}{1-\pageref{LastPage}}{10/24; Revised ?/?}{?/?}{21-0000}{Jhoan K. Hoyos-Osorio and Luis G. Sánchez-Giraldo}

\ShortHeadings{The Representation Jensen-Shannon Divergence}{Hoyos-Osorio and Sánchez-Giraldo}
\firstpageno{1}
\usepackage{caption}
\usepackage{tablefootnote}

\begin{document}
% The Jensen-Shannon Divergence from Covariance Operators on Reproducing Kernel Hilbert Spaces
\title{The Representation Jensen-Shannon Divergence}
% The Jensen-Shannon Divergence for Infinite Dimensional Covariance Operators in RKHS
\author{\name Jhoan K. Hoyos-Osorio \email keider.hoyos@uky.edu \\
       \addr Department of Electrical and Computer Engineering\\
       University of Kentucky\\
       Lexington, KY 40503, USA
       \AND
       \name Luis G. Sánchez-Giraldo \email luis.sanchez@uky.edu \\
       \addr Department of Electrical and Computer Engineering\\
       University of Kentucky\\
       Lexington, KY 40503, USA}

\editor{My editor}

\maketitle

\begin{abstract}%   <- trailing '%' for backward compatibility of .sty file
Quantifying the difference between probability distributions is crucial in machine learning. However, estimating statistical divergences from empirical samples is challenging due to unknown underlying distributions. This work proposes the representation Jensen-Shannon divergence (RJSD), a novel measure inspired by the traditional Jensen-Shannon divergence. Our approach embeds data into a reproducing kernel Hilbert space (RKHS), representing distributions through uncentered covariance operators. We then compute the Jensen-Shannon divergence between these operators, thereby establishing a proper divergence measure between probability distributions in the input space. We provide estimators based on kernel matrices and empirical covariance matrices using Fourier features. Theoretical analysis reveals that RJSD is a lower bound on the Jensen-Shannon divergence, enabling variational estimation. Additionally, we show that RJSD is a higher-order extension of the maximum mean discrepancy (MMD), providing a more sensitive measure of distributional differences. Our experimental results demonstrate RJSD's superiority in two-sample testing, distribution shift detection, and unsupervised domain adaptation, outperforming state-of-the-art techniques. RJSD's versatility and effectiveness make it a promising tool for machine learning research and applications.
\end{abstract}


\section{Introduction}

Divergences are functions that quantify the difference from one probability distribution to another. In machine learning, divergences can be applied to various tasks, including generative modeling (generative adversarial networks, variational auto-encoders), two-sample testing, anomaly detection, and distribution shift detection. The family of $f$-divergences is among the most popular statistical divergences, including the well-known Kullback-Leibler \citep{kullback1951information} and Jensen-Shannon divergences \citep{lin1991divergence}. A fundamental challenge to using divergences in practice is that the underlying distribution of data is unknown, and thus, divergences must be estimated from observations. Several divergence estimators have been proposed \citep{yang1999information,sriperumbudur2012empirical,krishnamurthy2014nonparametric,moon2014multivariate, singh2014generalized,li2016renyi,noshad2017direct, moon2018ensemble, bu2018estimation, berrett2019efficient, liang2019estimating, han2020optimal, sreekumar2022neural},  most of which fall into three categories: plug-in estimators, $k$-nearest neighbors, and neural estimators. 

 One alternative way of comparing distributions is by first mapping them to a representation space and then computing the distance between the mapped distributions. This approach is well-behaved if the mapping is injective, guaranteeing that different distributions are mapped to distinct points in the representation space. Dealing with the distributions in the new representation space can offer computational as well as statistical advantages (estimation from data). For example, the maximum mean discrepancy (MMD) \citep{gretton2012kernel} can be obtained by mapping the distributions into a reproducing kernel Hilbert space (RKHS) and computing the distance between embeddings. In this approach, distributions are mapped to what is called the mean embedding. In a similar vein, covariance operators (second-order moments) in RKHS have been used to propose distribution divergences \citep{harandi2014bregman, minh2015affine, zhang2019optimal, minh2021regularized, minh2023entropic}. Most of these divergences quantify the dissimilarity between Gaussian measures characterized by their respective covariance operators. However, the assumption of Gaussianity is not necessarily valid and might not effectively capture the disparity between the input distributions. 

Due to the underlying geometry, MMD lacks a straightforward connection with classical information theory tools \citep{bach2022information}. Alternatively, several information-theoretic measures based on kernel methods have been recently proposed to derive quantities that behave similarly to marginal, joint, and conditional entropy \citep{sanchezgiraldo2014measures, bach2022information}, as well as multivariate mutual information \citep{yu2019multivariate}, and total correlation \citep{yu2021measuring}. However, strategies for estimating divergences within this framework have been less explored. 

To fill this void, we propose a kernel-based information-theoretic framework for divergence estimation. We make the following contributions: 

\begin{itemize}
    \item We extend the Jensen-Shannon divergence between symmetric positive semidefinite matrices to infinite-dimensional covariance operators in reproducing kernel Hilbert spaces (RKHS). We show that this formulation defines a proper divergence between probability measures in the input space that we call the representation Jensen-Shannon divergence (RJSD).  
    
    \item RJSD avoids estimating the underlying density functions by mapping the data to an RKHS where distributions are embedded through uncentered covariance operators acting in this representation space. Notably, our formulation does not assume Gaussianity in the feature space. 
    
    \item We propose an estimator of RJSD from samples in the input space using Gram matrices. Consistency results for the proposed estimator are discussed.

    \item We established the connection between RJSD and the maximum mean discrepancy (MMD), demonstrating that MMD can be viewed as a particular case of RJSD.

        % \item We propose an estimator of RJSD from samples in the input space using Gram matrices. Additionally, we propose an estimator from empirical covariance matrices that explicitly maps data samples to an RKHS using Fourier features. \red{This estimator is flexible, scalable, differentiable, and suitable for minibatch-based optimization problems}. Consistency results and sample complexity bounds for the proposed estimators are discussed.
    
    \item The proposed divergence is connected to the classical Jensen-Shannon divergence of the underlying probability distributions. Namely, RJSD emerges as a lower bound on the classical Jensen-Shannon divergence, enabling the construction of a variational estimator.
\end{itemize}

% Different methods exist for representing data, including strings, graphs, lattices, statistical manifolds, etc. These representations are not necessarily vector spaces and therefore, traditional divergence estimators might not be applicable. In this case, we may only have access to data-dependent kernel functions that quantify the similarity between examples. Consequently, computing divergence between collections of objects based only on sample similarity is an appealing approach.

\subsection{Related Work}

Several divergences between covariance matrices in $\mathbb{R}^d$ have been extended to the infinite-dimensional covariance operators on reproducing kernel Hilbert spaces (RKHS) \citep{harandi2014bregman, minh2016covariance,minh2015affine, zhang2019optimal,quang2022kullback,minh2023entropic}.  In such cases, empirical estimation of the operators is handled implicitly using the kernel function associated with the RKHS. Thus, divergence computation uses the Gram matrix computed from pairwise evaluations of the kernel between data points.

Since covariance operators are Hilbert–Schmidt operators, the discrepancy between covariance operators can be measured by their Hilbert-Schmidt distance, which can be considered a generalization of the distance between covariance matrices induced by the Frobenius norm. For example, the Hilbert-Schmidt distance between empirical covariance operators admits a closed-form expression via the corresponding Gram matrices. If we use uncentered covariance operators, it can be shown that this distance is equivalent to the \emph{maximum mean discrepancy} (MMD) \citep{gretton2012kernel} with a squared kernel. Although this quantity has been widely used in the literature, the Hilbert–Schmidt distance disregards the manifold where the covariance operators live \citep{minh2016covariance}.

Some authors have applied the theory of symmetric positive definite matrices to measure the distance between potentially infinite-dimensional covariance operators while respecting the underlying geometry of these objects. 
These infinite-dimensional formulations are notably intricate, with regularization frequently proving necessary \citep{ha2014log,quang2022kullback,minh2023entropic}. Since logarithm, inverse, and determinant are typically involved in divergence/distance computation, regularization is required to ensure positive definiteness. For example, \citet{harandi2014bregman} extend some Bregman divergences for infinite dimensional covariance matrices (operators) in RKHS and provide closed-form expressions for Log determinant (Burg), and two symmetrized Bregman divergences, namely, the Jeffreys and Jensen-Bregman log determinant divergences. Similarly, \citet{ha2014log} investigates the estimation of the Log-Hilbert-Schmidt metric between covariance operators in RKHS, which generalizes the log-Euclidean metric. Later, \citet{minh2015affine} investigates the affine-invariant Riemannian distance between infinite-dimensional covariance operators and derives a closed-form expression to estimate it from Gram matrices. 

%One particular advantage of our proposed divergence compared to the above distances/divergences is that the finite and infinite-dimensional divergence formulas are the same, no regularization is necessary, and its closed-form estimator from Gram-matrices is straightforward. 

Some previously discussed divergences quantify the discrepancy between Gaussian measures characterized by their respective covariance operators. This framework assumes the data is distributed according to a Gaussian measure within the RKHS. This is the case of the log-determinant divergence, which corresponds to the Kullback-Leibler divergence between zero-mean Gaussian measures. Recently, \citet{minh2021regularized} and \citet{quang2022kullback} present a generalization of the Kullback-Leibler and Rényi divergences between Gaussian measures described by their mean embeddings and covariance operators on infinite-dimensional Hilbert Spaces. Similarly, \citet{zhang2019optimal} investigates the optimal transport problem between Gaussian measures on RKHS and proposes the kernel Wasserstein distance and the kernel Bures distance. Along the same lines, \citet{minh2023entropic} proposes an entropic regularization of the Wasserstein distance between Gaussian measures on RKHS. Although the artificial assumption that the data follows a Gaussian distribution in the RKHS facilitates the computation of these divergences, there is no guarantee that the data distribution in the feature space is indeed Gaussian. 

Recently, \citet{bach2022information} proposed the kernel Kullback-Leibler divergence. This divergence is formulated as the relative entropy of the distributions' uncentered covariance operators in RKHS. Although the paper discusses important theoretical properties of this divergence, its primary purpose is to serve as an intermediate step for deriving a measure of entropy. No empirical estimators for the divergence are introduced or discussed.

Our research proposes a novel approach: the representation (kernel) Jensen-Shannon divergence between two probability measures. Our divergence does not rely on the assumption of Gaussianity in the RKHS. Instead, the input distributions are directly mapped to uncentered covariance operators on RKHS, which characterize the distributions. Next, we compute the Jensen-Shannon divergence between these operators, also known as quantum Jensen-Shannon or Jensen-von Neumann divergence. Importantly, we demonstrate that this divergence can be readily estimated from data samples using Gram matrices derived from kernel evaluations between pairs of data points.  

% This research introduces a novel divergence measure: the representation (kernel) Jensen-Shannon divergence between two probability measures. Unlike prior approaches, our divergence does not assume that the data follows a Gaussian distribution in the RKHS. Instead, we directly map the input distributions to uncentered covariance operators in the RKHS, which characterize the distributions. We then compute the Jensen-Shannon divergence between these operators, extending the Jensen-von Neumann (quantum Jensen-Shannon) divergence between positive semidefinite matrices (density matrices) to covariance operators. Crucially, we establish that this divergence can be computed straightforwardly and in closed form using Gram matrices derived from pairwise kernel similarities in the input space. This approach simplifies the computation and circumvents the need for regularization, making it a robust and versatile tool for measuring discrepancies between complex data distributions.

% \red{motivation: Use ideas of kernel PCA: https://www.mlpack.org/papers/kpca.pdf}
\section{Preliminaries and Background}

In this section, we introduce the notation and discuss fundamental concepts. 

\subsection{Notation}
Let $(\mathcal{X}, \mathcal{F})$ be a measurable space. Let $\mathcal{M}_{+}^1(\mathcal{X})$ be the space of probability measures on $\mathcal{X}$, and let $P,$ $Q \in  \mathcal{M}_{+}^1(\mathcal{X})$ be two probability measures dominated by a $\sigma$-finite measure $\lambda$ on $(\mathcal{X}, \mathcal{F})$ (Similar notation from  \citet{stummer2012bregman}). Then, the densities $p = \frac{\d P}{d\lambda}$ and $q = \frac{\d Q}{d\lambda}$ have common support (the densities are positive on $\mathcal{X}$). $X \sim P$ and $Y \sim Q$ are two random variables distributed according to $P$ and $Q$. 

\subsection{Kernel Mean Embedding}
Let $\kappa : \mathcal{X} \times \mathcal{X} \rightarrow \mathbb{R}_{\geq 0}$ be a positive definite kernel. There exists a mapping $\phi: \mathcal{X} \rightarrow \mathcal{H}$, where $\mathcal{H}$ is a reproducing kernel Hilbert space, such that $\kappa(x, x') = \langle \phi(x),\phi(x')\rangle_{\mathcal{H}}$. The kernel mean embedding is a mapping $\mu$ from $\mathcal{M}_{+}^1(\mathcal{X})$ to $\mathcal{H}$ defined as follows \citep{smola2007hilbert}: 
For $P \in \mathcal{M}_{+}^1(\mathcal{X})$,
\begin{equation*}\label{eq:mean_embedding}
\mu_P = \mathbb{E}_{X\sim P}[\phi(X)] = \int\limits_{\mathcal{X}} \phi(x) \operatorname{d}P(x)
\end{equation*}

For a bounded kernel, $\kappa(x,x) < \infty$ for all $x \in \mathcal{X}$, we have that for any $f \in \mathcal{H}$, $\mathbb{E}_{X\sim P}[f(X)] = \langle f, \mu_{\scriptscriptstyle P} \rangle_{\mathcal{H}}$. 

% We can observe that Eqn. \ref{eq:mean_embedding} transforms the distribution $P$ to an element $\mu_{\mathbb
% {P}}$ in the feature space $\mathcal{H}$. This representation possesses several properties that can be exploited to characterize distributions when the kernel function is characteristic.
\subsection{Covariance Operator}

Another related mapping is the uncentered covariance operator \citep{baker1973joint}, one of the most important and widely used tools in RKHS theory. In this case, $P \in \mathcal{M}_{+}^1$ is mapped to an operator $\CP:\mathcal{H} \rightarrow \mathcal{H}$ given by:
\begin{equation}\label{eq:uncenetered_covariance_operator}
    \CP = \mathbb{E}_{X\sim P}[\phi(X)\otimes\phi(X)] = \int_\mathcal{X} \phi(x)\otimes\phi(x)\operatorname{d}P(x),
\end{equation} 
 where $\otimes$ is the tensor product. Similarly, for any $f,g \in \mathcal{H}$, $\mathbb{E}_{X\sim P}[f(X)g(X)] = \langle g, \CP f \rangle_{\mathcal{H}}$. 

The centered covariance operator is similarly defined as: 

\begin{equation*}
    \Sigma_{\scriptscriptstyle P} = \int_{\mathcal{X}}(\phi(X) - \mu_{\scriptscriptstyle P})\otimes (\phi(X) - \mu_{\scriptscriptstyle P})\operatorname{d}P(x) = \CP - \mu_{\scriptscriptstyle P}\otimes \mu_{\scriptscriptstyle P}.
\end{equation*}
% \subsubsection{Properties of the covariance operator}

The covariance operator is positive semidefinite and Hermitian (self-adjoint). Additionally, if the kernel is bounded, that is $\kappa(x,y) < \infty$, the covariance operator is trace class \citep{sanchezgiraldo2014measures,bach2022information}. Therefore, the spectrum of the covariance operator is discrete and consists of non-negative eigenvalues $\lambda_i$ with $\sum \lambda_i < \infty$, for which we can extend functions on $\mathbb{R}$ such as $t\log(t)$ and $t^\alpha$ to covariance operators via their spectrum \citep{Naoum2004functionalcalculuscompact}.

\subsection{Empirical Mean and Covariance}
Given $n$ samples $\mX = \{\vx_i\}_{i=1}^{n} \sim P$, the empirical mean embedding, and the empirical uncentered and centered covariance operators are defined as: 
\begin{equation*}\label{eq:empirical_mean_embedding}
    \muX = \frac{1}{n}\sum\limits_{i=1}^{n}\phi\left(\vx_i\right)
\end{equation*}
\begin{equation}\label{eq:empirical_uncentered_covariance_operator}
 \Cx = \frac{1}{n} \sum_{i=1}^n \phi(\vx_i)\otimes \phi(\vx_i)     
\end{equation}
\begin{equation*}\label{eq:empirical_centered_covariance_operator}
 \mSigma_{\scriptscriptstyle \mX} = \frac{1}{n} \sum_{i=1}^n (\phi(\vx_i) - \muX) \otimes (\phi(\vx_i) - \muX),
\end{equation*}

\section{Information Theory with Covariance Operators}
Throughout this paper, unless otherwise stated, we will assume that:
\\

\noindent \textbf{(A1)} $\kappa : \mathcal{X} \times \mathcal{X} \rightarrow \mathbb{R}_{\geq 0}$ is a positive definite kernel with an RKHS mapping $\phi: \mathcal{X} \rightarrow \mathcal{H}$ such that $\kappa(x, x') = \langle \phi(x),\phi(x')\rangle_{\mathcal{H}}$, and $\kappa(x, x) = 1 \quad \forall x\in \mathcal{X}$.
\\

Under this assumption, the covariance operator $\CP$ defined in Eqn.~\ref{eq:uncenetered_covariance_operator} is unit-trace. Note that since $\kappa(x,x) = 1$, we have that, $\Tr\left(\phi(x)\otimes \phi(x)\right) = \lVert \phi(x)\rVert^2 = 1$. Hence, the spectrum of the covariance operator consists of non-negative eigenvalues $\lambda_i$ with $\sum \lambda_i  = 1$, for which we can extend notions of entropy from the spectrum of unit-trace covariance operators. 
\begin{definition}
\label{def:rep_entropy}
    Let $X$ be a random variable taking values in $\mathcal{X}$ and probability measure $P$. Assume $\mathbf{(A1)}$ holds, and let $\CP$ be the corresponding unit-trace covariance operator defined in Eqn.~\ref{eq:uncenetered_covariance_operator}. Then, the representation (kernel) entropy of $X$ is defined as:
    \begin{align*}
  H^{\mathcal{H}}(X) = S(\CP) = -\Tr\left(\CP \log{\CP}\right),
\end{align*}
\end{definition}
  
\noindent where $S(\cdot)$ is a generalization of the von Neumann entropy \citep{von2018mathematical} for trace class operators, and it can be equivalently formulated as $S(\CP) = -\sum \lambda_i \log \lambda_i$. 

Similarly, the representation (kernel) Rényi entropy can be defined as \citep{sanchezgiraldo2014measures} : 

\begin{align*}
    H_\alpha^\mathcal{H}(X) = S_\alpha(\CP) = \frac{1}{1-\alpha} \log \biggl(\Tr\left(\CP^\alpha\right) \biggr), 
\end{align*}

\noindent where $\alpha>0$ is the entropy order. Notice that in the limit when $\alpha \rightarrow 1$,  $H^\mathcal{H}_{\alpha \rightarrow 1} =  H^\mathcal{H}(X)$. These quantities resemble the quantum von-Neumann and quantum Rényi entropy \citep{muller2013quantum} where the covariance operator plays the role of a density matrix. Although the representation entropy has similar properties to those of Shannon (or Rényi) entropy, it is important to emphasize that the representation entropy is not equivalent to these entropies, and thus estimating representation entropy does not amount to estimating Shannon or Rényi entropies. Instead, the representation entropy incorporates the data representation. Its properties are not only determined by the data distribution but also depend on the representation (kernel). 

\subsection{Empirical Estimation of Representation Entropy}
\label{sec:kernel_itl}
Let $\mX = \{\vx_i\}_{i=1}^{n} \sim P$ be $n$ i.i.d samples of a random variable $X$ with probability measure $P$. An empirical estimate of representation entropy can be obtained based on the spectrum of the empirical uncentered covariance operator $\Cx$ defined in Eqn. \ref{eq:empirical_uncentered_covariance_operator}. Consider the Gram matrix $\Kx$, consisting of all pairwise kernel evaluations between data points in the sample $\mX$, that is, $(\Kx)_{ij} = \kappa(\vx_i, \vx_j)$ for $i, j = 1, \dots, n$.  It can be shown that $\Cx$ and $\frac{1}{n}\Kx$ have the same non-zero eigenvalues \citep{sanchezgiraldo2014measures, bach2022information}. Based on this equivalence, the estimator of representation entropy can be expressed in terms of the Gram matrix $\Kx$ as follows:
\begin{proposition}
    The empirical kernel-based representation entropy estimator of $X$ is
\begin{equation}\label{eq:matrix_based_entropy}
\hat{H}^\mathcal{H}(X) = S(\Cx) = S\left(\tfrac{1}{n}\Kx\right)= -\Tr{\left(\tfrac{1}{n}\Kx\log \tfrac{1}{n}\Kx\right) } = -\sum_{i=1}^{n}\lambda_i\log\lambda_i,
\end{equation}
\end{proposition}

\noindent where $\lambda_i$ denotes the $i$th eigenvalue of $\tfrac{1}{n}\Kx$. The eigen-decomposition of $\Kx$ has $\mathcal{O}(n^3)$ time complexity. Next, we show the estimation bounds for the representation entropy estimator, which converges to the population quantity at a rate of $\mathcal{O}(1/\sqrt{n})$:

\begin{proposition}
\label{prop:convergence}
    \citep{bach2022information}[Proposition 7] Assume that $P$ has a density with respect to the uniform measure that is greater than $\alpha < 1$. Finally, assume that $c = \int_0^\infty \sup_{x\in \mathcal{X}}\langle\phi(x),(\CP + \lambda I)^{-1}\phi(x)\rangle^2d\lambda$ is finite. Then: 

    \begin{equation*}
        \mathbb{E}\left[S(\tfrac{1}{n}\Kx) - S(\CP) \right] \leq \frac{1+c(8\log(n))^2}{n} + \frac{17}{\sqrt{n}}(2\sqrt{c} + \log(n)). 
    \end{equation*}
\end{proposition}

% \paragraph{Covariance-based estimator:} 

This estimator of kernel-based representation entropy can be used in gradient-based learning \citep{sanchez2013iclr, sriperumbudur2015optimal}. Representation entropy has been used as a building block for other matrix-based measures, such as joint and conditional representation entropy, mutual information \citep{yu2019multivariate}, total correlation \citep{yu2021measuring}, and divergence \citep{osorio2022representation, bach2022information}. 

\section{The Representation Jensen-Shannon Divergence}
For two probability measures $P$ and $Q$ on a measurable space $(\mathcal{X},\mathcal{F})$, the Jensen-Shannon divergence (JSD) is defined as follows:
\begin{equation*}
    D_{\scriptscriptstyle JS}(P,Q) = H\left(\frac{P + Q}{2}\right) - \frac{1}{2}\left(H(P) + H(Q)\right),
\end{equation*}
where $\frac{P + Q}{2}$ is the mixture of $P$ and $Q$ and $H(\cdot)$ is Shannon's entropy. Properties of JSD, such as boundedness, convexity, and symmetry, have been extensively studied \citep{briet2009properties, sra2021metrics}.
The Quantum counterpart of the Jensen-Shannon divergence (QJSD) between density matrices \footnote{A density matrix is a unit-trace symmetric positive semidefinite matrix that describes the quantum state of a physical system}  $\rho$ and $\sigma$ is defined as  $D_{\scriptscriptstyle JS}(\rho,\sigma)  = S\left(\frac{\rho + \sigma}{2}\right) - \frac{1}{2}\left(S(\rho) + S(\sigma)\right)$, where $S(\cdot)$ is von Neumann's entropy. QJSD is everywhere defined, bounded, symmetric, and positive if $\rho \neq  \sigma$ \citep{sra2021metrics}. Like the kernel-based entropy where the uncentered covariance operator is used in place of a density matrix, we derive a measure of divergence where $\rho$ and $\sigma$  are replaced by the uncentered covariance operators corresponding to $P$ and $Q$. 
\label{sec:matrix_based JSD}
\begin{definition}
\label{def:rep_JSD}
    Let $P$ and $Q$ be two probability measures defined on a measurable space $(\mathcal{X},\mathcal{F})$, and $\mathbf{(A1)}$ is satisfied. Then, the \textbf{representation Jensen-Shannon divergence} (RJSD) between $P$ and $Q$ is defined as:  
\begin{equation*}\label{eq:representation_JSD}
    D_{\scriptscriptstyle JS}^\mathcal{H}(P,Q) = D_{\scriptscriptstyle JS}(\CP,\CQ) = S\left(\frac{\CP + \CQ}{2}\right) - \frac{1}{2}\left(S(\CP) + S(\CQ)\right).
\end{equation*}
\end{definition}

\subsection{Properties}
First, we show that RJSD relates to the maximum mean discrepancy (MMD) with kernel $\kappa^2$, where MMD is defined as $\operatorname{MMD}^2_\kappa(P, Q) = \lVert  \mu_{\scriptscriptstyle P} - \mu_{\scriptscriptstyle Q} \rVert_\mathcal{H}^2$.  

\begin{lemma} \label{th:upper_bound}
For all probability measures $P$ and $Q$ defined on $\mathcal{X}$, and covariance operators $\CP$ and $\CQ$ with RKHS mapping $\phi(x)$ such that $\langle \phi(x),\phi(x)\rangle_{\mathcal{H}} = 1 \quad \forall x \in \mathcal{X}$: 
\begin{equation*}
    D_{\scriptscriptstyle JS}^{\mathcal{H}}(P,Q) \geq \frac{1}{8} \lVert C_{\scriptscriptstyle P} - C_{\scriptscriptstyle Q}\rVert_{\scriptscriptstyle HS}^2 = \frac{1}{8} \operatorname{MMD}^2_{\kappa^2} (P,Q)
\end{equation*}  
Proof: \emph{See Appendix \ref{ap:proof_upper_bound}.}
\end{lemma}

\begin{theorem}
   Let $\kappa^2$ be a characteristic kernel. Then, the representation Jensen-Shannon  divergence  $D_{\scriptscriptstyle JS}^{\mathcal{H}}(P,Q) = 0$ if and only if $P=Q$.   
\end{theorem}
\begin{proof}
    It is clear that if $P=Q$ then $D_{\scriptscriptstyle JS}^{\mathcal{H}}(P,Q) = 0$. We now prove the opposite. According to Lemma \ref{th:upper_bound}, $D_{\scriptscriptstyle JS}^{\mathcal{H}}(P,Q) = 0$ implies that $\operatorname{MMD}^2_{\kappa^2} (P,Q) = 0$. Then, if $\operatorname{MMD}^2_{\kappa^2} (P,Q) = 0$ and the kernel $\kappa^2$ is characteristic, then $P = Q$ \citep{gretton2012kernel}, completing the proof.  
\end{proof}

This theorem demonstrates that RJSD defines a proper divergence between probability measures in the input space. In summary, RJSD inherits most of the classical and quantum Jensen-Shannon divergence properties. 
\begin{itemize}
    \item \emph{Non-negativity}: $ D_{\scriptscriptstyle JS}^\mathcal{H}(P, Q) \geq 0 $.
    \item \emph{Positivity:} $ D_{\scriptscriptstyle JS}(\CP, \CQ) = 0$ if and only if $\CP =  \CQ$. If the kernel $\kappa^2$ is characteristic, $D_{\scriptscriptstyle JS}^\mathcal{H}(P, Q) = 0$ if and only if $P = Q$. 
    \item \emph{Symmetry:} $    D_{\scriptscriptstyle JS}^\mathcal{H}(P, Q) =  D_{\scriptscriptstyle JS}^\mathcal{H}(Q, P)$.
    \item \emph{Boundedness:}     $ D_{\scriptscriptstyle JS}^\mathcal{H}(P, Q) \leq \log (2)$,
    \item    $ D_{\scriptscriptstyle JS}(\CP, \CQ)^{\frac{1}{2}}$ is a metric on the cone of uncentered covariance matrices in any dimension \citep{virosztek2021metric}. 
\end{itemize}

Additionally, we introduce a fundamental property of RJSD and its connection with its classical counterpart.

\begin{theorem} \label{th:lower_bound}
For all probability measures $P$ and $Q$ defined on $\mathcal{X}$, and unit-trace covariance operators $\CP$ and $\CQ$, the following inequality holds: 
\begin{equation}\label{eq:lower_bound}
    D_{\scriptscriptstyle JS}^\mathcal{H}(P, Q) \leq D_{\scriptscriptstyle JS}(P, Q)
\end{equation} 

Proof: \emph{See Appendix \ref{ap:proof_lower_bound}.}
\end{theorem}
% \begin{theorem} \label{th:divergence_equality}
% \red{let $P$ and $Q$ be two probability measures defined on $\mathcal{X}$, with probability density functions $p$ and $q$ respectively. If there exists a mapping $\phi^*$ such that $ p(x) = \frac{1}{h_P} \left\langle \phi^*(x), \CP \phi^*(x) \right\rangle $ and $q(x) = \frac{1}{h_Q} \left\langle \phi^*(x), \CQ \phi^*(x) \right\rangle $, then:}
% \begin{align*}
% D_{\scriptscriptstyle JS}^{\mathcal{H}}(P,Q) & = D_{\scriptscriptstyle JS}(P,Q).   
% \end{align*}
% Proof: \emph{See Appendix \ref{ap:proof_equality}.}
% \end{theorem}
% \red{This theorem implies that the bound in Eqn. \ref{eq:lower_bound} is tight for optimal functions $\phi(x)$
% that approximate the true underlying distributions through Eqn. \ref{eq:kernel_density}}. 
Theorem \ref{th:lower_bound} can be used to obtain a variational estimator of Jensen-Shannon divergence (see Section \ref{sec:variational_estimation}).

% \red{The result of Theorem \ref{th:upper_bound} should not be underestimated. Since MMD is a lower bound on the RJSD, any discrepancies between distributions that can be detected with MMD should be also detected with RJSD. That is RJSD should be at least as good as MMD. Moreover, it also shows that RJSD is well defined for characteristic kernel, for which RJSD is non zero if $P \neq Q$. } 

\subsection{Empirical Estimation of the Representation Jensen-Shannon Divergence}
Given two sets of samples $\mX = \left\{\vx_i\right\}_{i=1}^n \subset \mathcal{X}$ and $\mY = \left\{\vy_i\right\}_{i=1}^m \subset \mathcal{X}$ drawn from two unknown probability measures $P$ and $Q$, we propose the following RJSD estimator: 
\paragraph{Kernel-based estimator:} Let $\kappa$ be a positive definite kernel, $\mZ$ be the mixture of the samples of $\mX$ and $\mY$, that is, $\mZ = \left\{\vz_i\right\}_ {i=1}^{n+m}$ where $\vz_i = \vx_i$ for $i\in \{1,\dots,n\}$ and $\vz_i = \vy_{i-n}$ for $i\in \{n+1, \dots, n+m\}$. Finally, let $\Kz$ be the kernel matrix consisting of all normalized pairwise kernel evaluations of the samples in $\mZ$, that is, the samples from both distributions. Moreover, let $\Kx$ and $\Ky$ be the pairwise kernel matrices of $\mX$ and $\mY$ respectively.

Notice that the sum of uncentered covariance operators in the RKHS corresponds to the covariance operator of the mixture of samples in the input space, that is, $\tfrac{n}{n+m}\Cx + \tfrac{m}{n+m}\Cy = \Cz $.

% \begin{align*}
%     \tfrac{n}{n+m}\Cx + \tfrac{m}{n+m}\Cy &= \tfrac{1}{n+m}\mPhiX^{\top}\mPhiX + \tfrac{1}{n+m}\mPhiY^{\top}\mPhiY\\
%     & = \tfrac{1}{n+m}\begin{bmatrix}
%  \mPhiX^{\top} & \mPhiY^{\top}
% \end{bmatrix}
% \begin{bmatrix}
%  \mPhiX\\
%  \mPhiY
% \end{bmatrix} \\
% & = \tfrac{1}{n+m}\mPhiZ^{\top}\mPhiZ = \Cz,
% \end{align*}

Since $\Cz, \Cx, \Cy$ and $\tfrac{1}{n+m}\Kz,\tfrac{1}{n}\Kx,\tfrac{1}{m}\Ky$ share the same non-zero eigenvalues respectively, the divergence can be directly computed from samples in the input space as follows.
\begin{proposition} 
The empirical kernel-based RJSD estimator for a kernel $\kappa$ is
   \begin{equation}\label{eq:JSD_kernel}
    \widehat{D}_{\scriptscriptstyle JS}^{\: \kappa}(\mX,\mY) = S\left(\tfrac{1}{n+m}\Kz\right) - \left(\tfrac{n}{n+m}S\left(\tfrac{1}{n}\Kx\right) + \tfrac{m}{n+m}S\left(\tfrac{1}{m}\Ky\right)\right).
\end{equation}   
\end{proposition}
  
Leveraging the convergence results in Proposition \ref{prop:convergence}, we can show that  $\widehat{D}_{\scriptscriptstyle JS}^{\: \kappa}(\mX, \mY)$ converges to the population quantity at a rate $\mathcal{O}\left(\frac{1}{\sqrt{n}}\right)$, assuming $n=m$ (Appendix \ref{ap:proof_convergence_kernel}). However, notice that $S\left(\tfrac{1}{n+m}\Kz\right)$ converges faster to $S(\Cz)$ than $S\left(\tfrac{1}{n}\Kx\right)$ and $S\left(\tfrac{1}{m}\Ky\right)$ to $S(\Cx)$ and $S(\Cy)$ respectively. This faster convergence is because we use more samples ($n+m$) to estimate $S(\Cz)$ than $S(\Cx)$ and $S(\Cy)$. This imbalance allows $S(\Cz) \leq \log(n+m)$ to estimate up to larger entropy values compared to $S(\Cx) \leq \log(n)$ and $S(\Cy)\leq \log(m)$. Therefore, the estimator in Eqn. \ref{eq:JSD_kernel} exhibits an upward bias. Next, we propose an alternative estimator to reduce this effect.

% \red{Additionally, a direct consequence of Theorem \ref{th:convergence_1} is that under the same assumptions of the theorem, $D_{\scriptscriptstyle JS}^{\kappa}(\mX,\mY)$ converges to  $D_{\scriptscriptstyle JS}(P,Q)$ as $n\rightarrow \infty$ with probability one. }    

\subsubsection{Addressing the Upward Bias of the Kernel-based Estimator}
The upward bias described above causes an undesired effect in the divergence. The kernel RJSD estimator can be trivially maximized when the sample's similarities are negligible, for example, when the kernel bandwidth $\sigma$ in a Gaussian kernel is close to zero (see Fig. \ref{fig:RJSD_varying_samples}). This behavior is caused by the discrepancy between the number of samples used to estimate $S(\tfrac{1}{n+m}\Kz)$ compared to $S(\tfrac{1}{n}\Kx),$ and $S(\tfrac{1}{m}\Ky)$, which causes $S(\tfrac{1}{n+m}\Kz)$ to grow faster and up to $\log(n+m)$ compared to $S(\tfrac{1}{n}\Kx)$ and $S(\tfrac{1}{m}\Ky)$ that can only grow up to $\log(n)$ and $\log(m)$ respectively (see Fig. \ref{fig:approximation}, rightmost). To reduce the bias of the estimator in Eqn. \ref{eq:JSD_kernel} and avoid trivial maximization, we need to regularize $S(\tfrac{1}{n+m}\Kz)$ so that it estimates up to similar values of entropy than $S(\tfrac{1}{n}\Kx)$ and $S(\tfrac{1}{m}\Ky)$. We propose the following alternatives:

\paragraph{Power Series Expansion Approximation: }

% Another mechanism to regularize the divergence is by using a power series expansion of the logarithm of the matrix.

Let $\mA$ be a positive semidefinite matrix, such that $\lVert \mA \rVert_2 \leq 1$, where $\lVert\mA\rVert_2 = \max_i(\lambda_i)$ denotes the spectral or $\normltwo$-norm, (which is the case for all trace-normalized kernel matrices). Then, the following power series expansion converges to $\log(\mA)$ \citep{higham2008functions}: 

\begin{equation*}
    \log(\mA) = - \sum_{j=1}^\infty \frac{(\mI - \mA)^j }{j}.    
\end{equation*}

We propose approximating the logarithm by truncating this series to a lower order. 

\begin{proposition}
The power-series kernel entropy estimator of $X$ is:
    \begin{equation*}
            S_p(\tfrac{1}{n}\Kx)=\sum_{j=1}^{p}\frac{1}{j} \Tr \left(\tfrac{1}{n}\Kx \left(\mI - \tfrac{1}{n}\Kx \right)^j \right), 
    \end{equation*}
\noindent where $p$ is the order of the approximation. 
\end{proposition}

\begin{proposition} 
The power-series RJSD estimator is
   \begin{equation*}\label{eq:JSD_power}
    \widehat{D}_{\scriptscriptstyle pJS}^{\: \kappa}(\mX,\mY) = S_p\left(\tfrac{1}{n+m}\Kz\right) - \left(\tfrac{n}{n+m}S_p\left(\tfrac{1}{n}\Kx\right) + \tfrac{m}{n+m}S_p\left(\tfrac{1}{m}\Ky\right)\right).
\end{equation*}   
\end{proposition}

This approximation has two purposes. First, it avoids the need for eigenvalue decomposition.
Second, it indirectly regularizes the three entropy terms of the divergence, where $\Kz$ is regularized more strongly due to its larger size. For example, $S_p(\Kz)\leq \sum\limits_{j=1}^{p}\tfrac{1}{j}(1-\tfrac{1}{n+m})^j$ while $S_p(\Kx)\leq \sum\limits_{j=1}^{p}\tfrac{1}{j}(1-\tfrac{1}{n})^j$ and $S_p(\Ky)\leq \sum\limits_{j=1}^{p}\tfrac{1}{j}(1-\tfrac{1}{m})^j$. 

By increasing the order, the gap between the maximum entropies obtained by the three entropy terms grows, leading to the behavior discussed above. Truncating the power series helps avoid trivial maximization of the divergence at lower kernel bandwidths (see Fig. \ref{fig:power_series}) or equivalently in high dimensions where some similarities could be insignificant and the kernel matrices could be sparse (see Fig. \ref{fig:power_series_dimension}). Consequently, the RJSD power series expansion offers a more robust estimator that goes beyond reducing computational costs.

Next, we show an important connection between the power-series RJSD estimator and MMD: 

\begin{theorem}
    Assume\textbf{(A1)} and let $p=1$ be the order of the power series expansion approximation. Then, given two sets of samples $\mX = \left\{\vx_i\right\}_{i=1}^n \sim P$ and $\mY = \left\{\vy_i\right\}_{i=1}^n  \sim Q$: 

    \begin{equation*}\label{eq:upper_bound}
    \widehat{D}_{\scriptscriptstyle pJS}^{\: \kappa}(\mX, \mY) =  \frac{1}{4} \widehat{\operatorname{MMD}}^2_{\kappa^2} (\mX,\mY)
\end{equation*}  
\end{theorem}
\begin{proof}
    \begin{align*}
    \widehat{D}_{\scriptscriptstyle pJS}^{\: \kappa}(\mX, \mY) &= \Tr\left(\tfrac{1}{2n}\Kz(\mI -\tfrac{1}{2n}\Kz )\right) - \frac{1}{2}\Tr\left(\tfrac{1}{n}\Kx(\mI -\tfrac{1}{n}\Kx )\right) - \frac{1}{2}\Tr\left(\tfrac{1}{n}\Ky(\mI -\tfrac{1}{n}\Ky )\right)\\
    & = -\Tr\left(\tfrac{1}{4n^2}\Kz\Kz\right) +\frac{1}{2}\Tr\left(\tfrac{1}{n^2}\Kx\Kx\right) + \frac{1}{2}\Tr\left(\tfrac{1}{n^2}\Ky\Ky\right)\\
    & = -\frac{1}{4n^2}\lVert\Kz\rVert_F^2 +\frac{1}{2n^2}\lVert\Kx\rVert_F^2 + \frac{1}{2n^2}\lVert\Ky\rVert_F^2 \\ 
    & = -\frac{1}{4n^2}\sum_{i,j}^{2n}\kappa^2(\vz_i,\vz_j) +\frac{1}{2n^2}\sum_{i,j}^{n}\kappa^2(\vx_i,\vx_j) + \frac{1}{2n^2}\sum_{i,j}^{n}\kappa^2(\vy_i,\vy_j) \\ 
    &= \frac{1}{4n^2}\sum_{i,j}^{n}\kappa^2(\vx_i,\vx_j) + \frac{1}{4n^2}\sum_{i,j}^{n}\kappa^2(\vy_i,\vy_j) - \frac{2}{4n^2}\sum_{i,j}^{n}\kappa^2(\vx_i,\vy_j)\\ 
    & =\frac{1}{4} \widehat{\operatorname{MMD}}^2_{\kappa^2} (\mX,\mY)
    \end{align*}
\end{proof}

This theorem establishes that RJSD extends MMD to higher-order statistics of the kernel matrices and the covariance operator. While MMD captures second-order interactions of data projected in the reproducing kernel Hilbert space (RKHS) defined by the kernel function $\kappa$, RJSD incorporates higher-order statistics, enhancing the measures' sensitivity to subtle distributional differences.

\paragraph{Finite-dimensional feature representation:} Next, we propose an alternative estimator using an explicit finite-dimensional feature representation based on Fourier features. For $\mathcal{X} \subseteq \mathbb{R}^d$ and a shift-invariant kernel $\kappa(x,x^{\prime}) = \kappa(x-x^{\prime})$, the random Fourier features (RFF) \citep{rahimi2007random} is a method to create a smooth feature mapping $\phi_{\omega}(x):  \mathcal{X} \to \Real^{2D}$ so that $\kappa(x-x') \approx \langle\phi_{\omega}(x), \phi_{\omega} (x')\rangle$. 

%

% \red{Use this notation for RFFs: https://arxiv.org/pdf/2006.05013.pdf}

For some data $\mX \in \Real^{n\times d}$, and kernel $\kappa(\cdot,\cdot)$ with Fourier transform $p(\omega)$, the corresponding random Fourier features $\mPhiXhat \in \Real^{n\times 2D}$ are obtained by computing $\mX\mW \in \Real^{n\times D}$, where $\mW \in \Real^{d\times D}$ is a random matrix such that each column of $\mW$ denoted by $\mW_{:j}$ is sampled from $p(\omega)$. Then point-wise cosine and sine nonlinearities are applied, that is,
\begin{equation*}
    \mPhiXhat = \sqrt{\frac{1}{D}} \begin{bmatrix}
        \cos(\mX\mW) & \sin(\mX\mW)
    \end{bmatrix} 
\end{equation*}
% Letting $\mPhiXhat = \left[ \phi_{\omega}(\vect{x}_1)^{T}, \phi_{\omega}(\vect{x}_2)^{T}, \cdots, \phi_{\omega}(\vect{x}_n)^{T} \right]^{T} $ be the $n\times D$ matrix containing the mapped samples, 

Let $\mPhiXhat \in \Real^{n\times 2D}$ and $\mPhiYhat \in \Real^{m\times 2D}$ be the matrices containing the mapped samples of $\mX$ and $\mY$. Then, the empirical uncentered covariance matrices are computed as $\hatCx=\frac{1}{n}\mPhiXhat^{\top}\mPhiXhat$ and $\hatCy=\frac{1}{m}\mPhiYhat^{\top}\mPhiYhat$. We propose the following covariance-based RJSD estimator. 

\begin{proposition}
The Fourier Features-based estimator is defined as:  
    \begin{align*} \label{eq:JSD_covariances}
    \widehat{D}_{\scriptscriptstyle fJS}^{\: \kappa}(\mX, \mY;\omega) &= S\left(\tfrac{n}{n+m}\hatCx + \tfrac{m}{n+m}\hatCy\right) - \left(\tfrac{n}{n+m}S(\hatCx) + \tfrac{m}{n+m}S(\hatCy)\right), 
    % \\ & = S\left(\pi_1\Cx + \pi_2\Cy\right) - \left(\pi_1S(\Cx) + \pi_2S(\Cy)\right)
\end{align*}
\end{proposition}

Using Fourier features to estimate RJSD offers additional benefits beyond reducing the computational burden. First, notice that by using explicit empirical covariance matrices, in the case of $2D < n,m$, the term $S\left(\tfrac{n}{n+m}\hatCx + \tfrac{m}{n+m}\hatCy\right)\leq \log(2D)$, likewise $S(\hatCx)$ and $S(\hatCy)$, which reduces the bias problem due to rank differences between the matrices. Additionally, the Fourier features allow parameterizing the representation space, which can be helpful for kernel learning. Accordingly, we can treat the Fourier features as learnable parameters within a neural network (Fourier features network), optimizing them to maximize divergence and enhance its discriminatory power. Finally, we can consider incremental updates to the covariance operators which can reduce the variance of the divergence estiamates when using minibatches. Consequently, the Fourier features approach offers a more versatile estimator that extends beyond reducing the computational cost.

\section{Experiments}
\subsection{Analyzing Estimator properties}

In this section, we study the behavior of the proposed estimators under different conditions. First, we analyze empirically the convergence of the kernel-based estimator as the number of samples increases. Here, $P(x; l_p, s_p)$ and $Q(x; l_q, s_q)$ represent two Cauchy distributions with location parameters $l_p$ and $l_q$, and scale parameters $s_p = s_q = 1$. To examine the relationship between the true JSD and the proposed estimators, we utilize the closed form of the JSD between Cauchy distributions derived by \citet{nielsen2022f}. According to \citet{bach2022information}, when the kernel bandwidth approaches zero and the number of samples $n$ approaches infinity, the kernel-based entropy converges to the classical Shannon entropy. Consequently, we expect the RJSD to be equivalent to the classical JSD at this limit.

% \begin{figure}
%     % \subfigure[]{
%     % \includegraphics[width=0.35\textwidth]{Figures/RJSD_samples.pdf}
%     % \label{fig:RJSD_varying_samples}}
%     % % \subfigure[]{
%     % % \includegraphics[width=0.3\textwidth]{Figures/RJSD_low_rank_samples.pdf}}
%     % \subfigure[]{
%     % \includegraphics[width=0.35\textwidth]{Figures/RJSD_power_series.pdf}
%     % \label{fig:power_series}}
%     % \subfigure[]{
%     % \includegraphics[width=0.35\textwidth]{Figures/tikhonov.pdf}}
%     % \subfigure[]{
%     % \includegraphics[width=0.35\textwidth]{Figures/RJSD_fourier_features.pdf}}
%     % % \subfigure[]{
%     % % \includegraphics[width=0.3\textwidth]{Figures/RJSD_estimators.pdf}}
%     \centering
%     \includegraphics[width=1.0\textwidth]{Figures/RJSD_estimators2.pdf}
%     \caption{RJSD estimators comparison with Gaussian kernel for two Cauchy distributions with Jensen-Shannon divergence $0.5\times \log(2)$, $n=256$ unless otherwise specified.}
%     \label{fig:RJSD estimators}
% \end{figure}

\begin{figure}[!t]
\subfigure[]{
\includegraphics[trim={0 0.4cm 0 0.1cm},clip, width=0.345\textwidth]{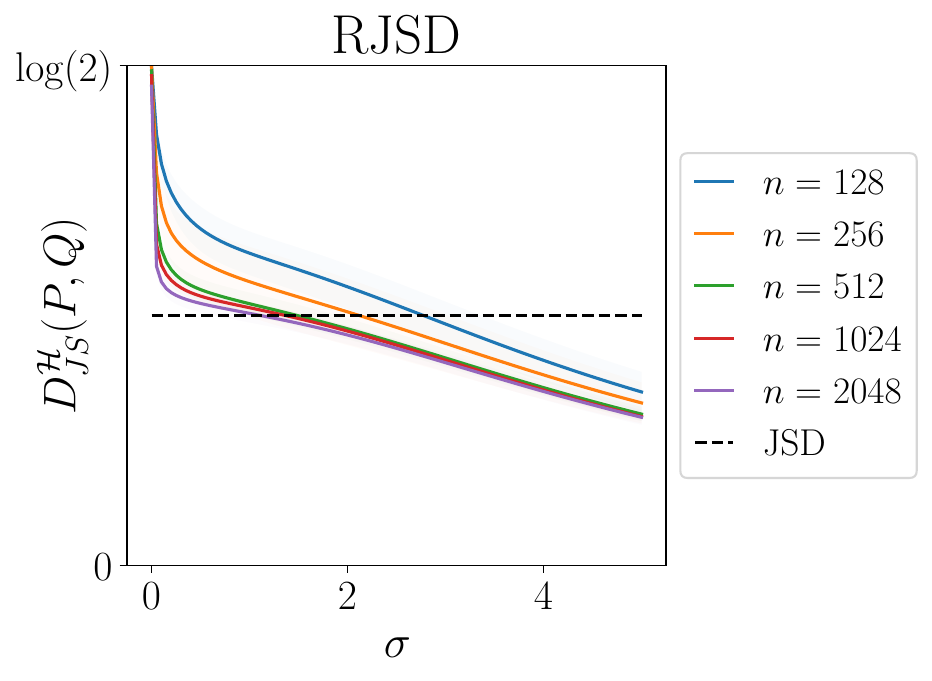}
\label{fig:RJSD_varying_samples}} 
\hspace{-0.6cm} 
\vspace{-0.1cm}
\subfigure[]{
\includegraphics[trim={0 0.35cm 0 0.1cm},clip, width=0.31\textwidth]{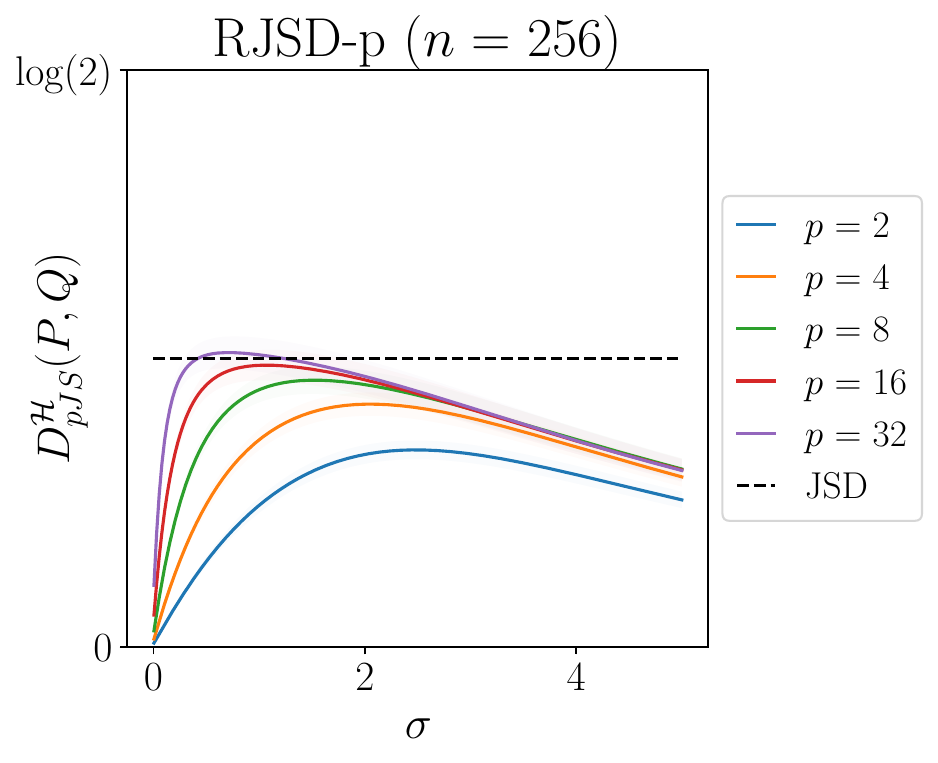}
\label{fig:power_series}}
\hspace{-0.5cm} 
\vspace{-0.1cm}
\subfigure[]{
\includegraphics[trim={0 0.35cm 0 0.1cm},clip, width=0.33\textwidth]{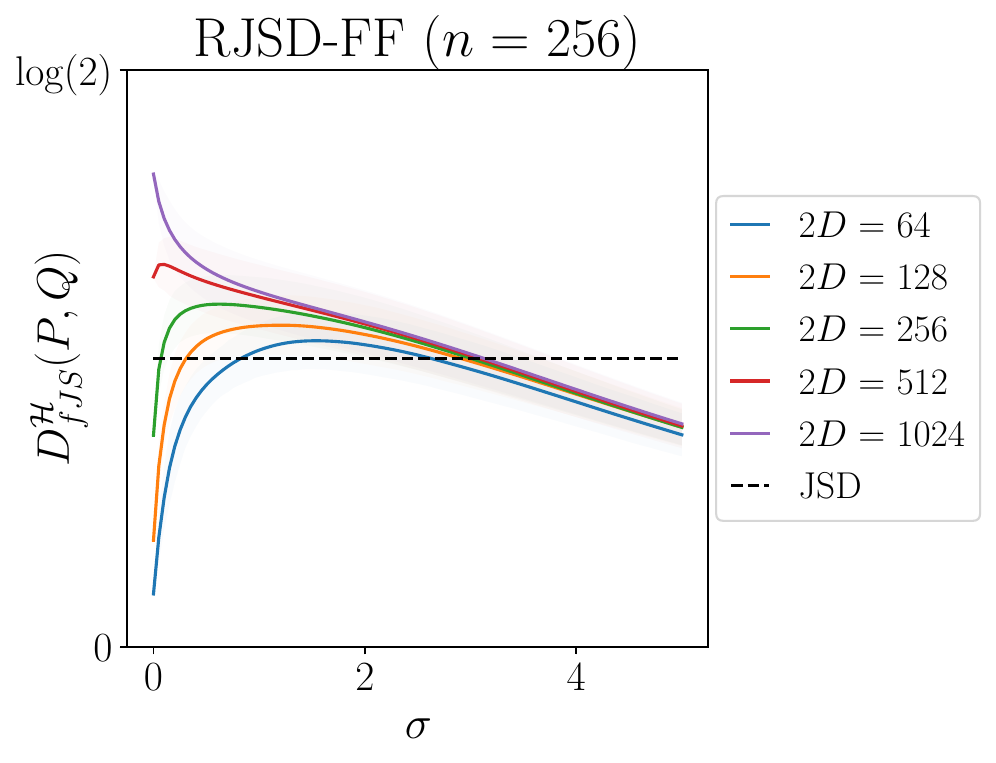}
\label{fig:fourier_features}}
\subfigure[]{
    \includegraphics[trim={0 0.35cm 0 0.1cm},clip, width=0.33\textwidth]{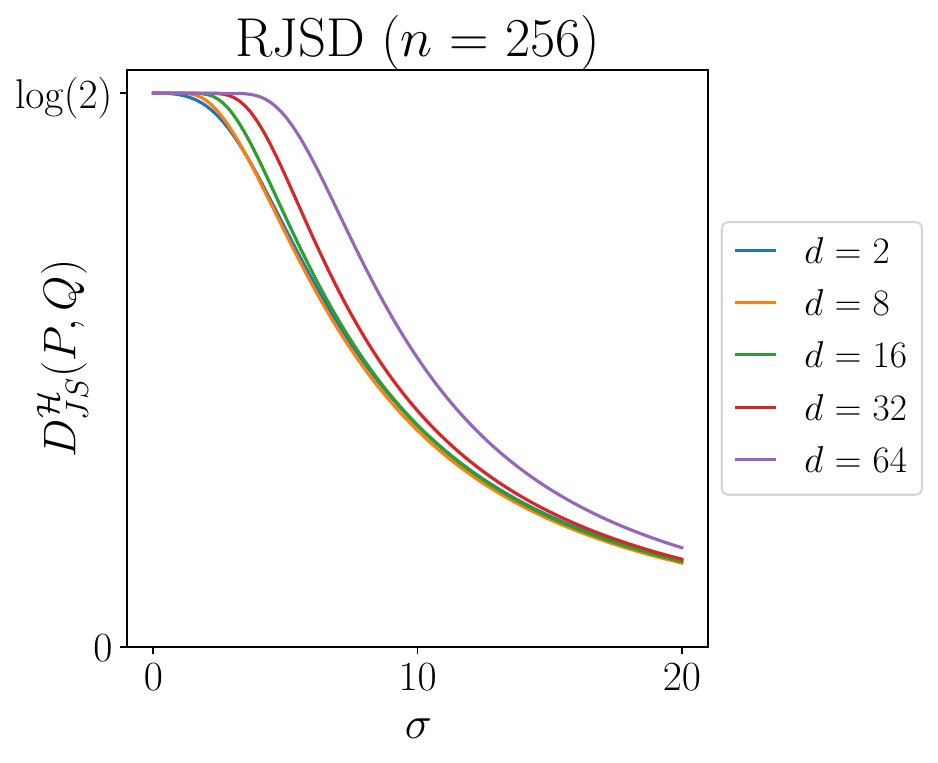}
    \label{fig:RJSD_dimensionality}}
    \hspace{-0.5cm}
    \vspace{-0.1cm}
    \subfigure[]{
    \includegraphics[trim={0 0.35cm 0 0.1cm},clip, width=0.33\textwidth]{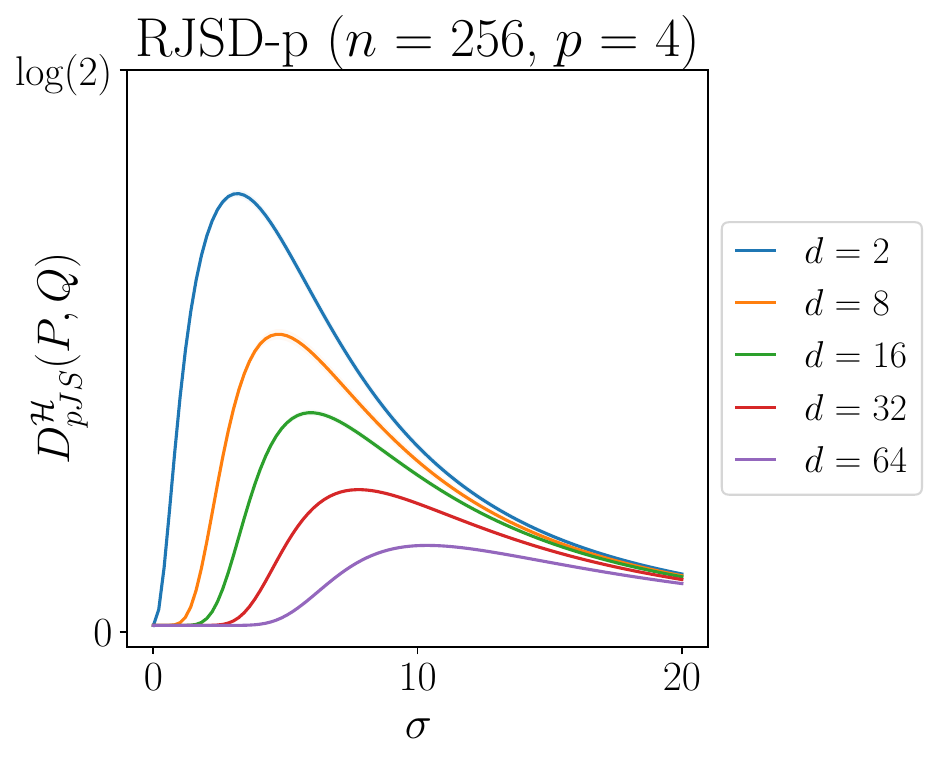}\label{fig:power_series_dimension}}
    \hspace{-0.35cm}
    \subfigure[]{
    \includegraphics[trim={0 0.35cm 0 0.1cm},clip, width=0.33\textwidth]{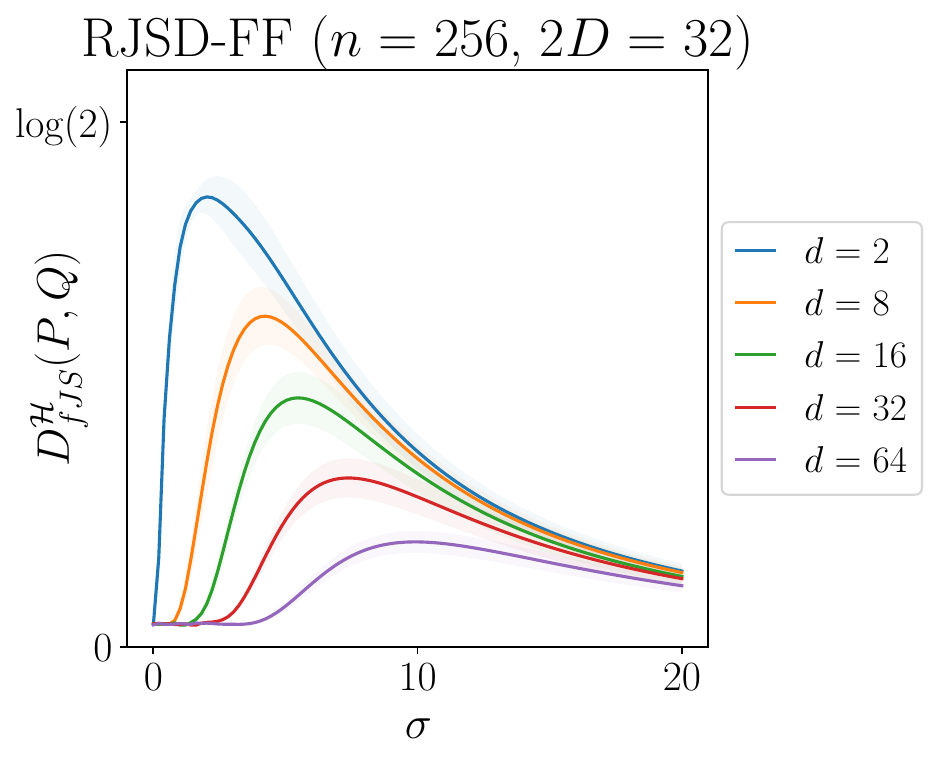}
    \label{fig:fourier_features_dimension}}
\caption{Comparing RJSD estimators with Gaussian kernel while varying the kernel bandwidth. The first row illustrates the divergence between two Cauchy distributions ($d=1$) with Jensen-Shannon divergence (JSD) $JSD = 0.5\times \log(2)$. The second row presents the estimated divergence for two multivariate Gaussians while varying dimensionality.}
\label{fig:RJSD estimators}
\end{figure}

Fig. \ref{fig:RJSD_varying_samples} illustrates the behavior of the kernel-based estimator using a Gaussian kernel, $\kappa(x,x') = \exp\left(-\frac{\lVert x-x' \rVert^2}{2\sigma^2}\right)$, as we vary both the bandwidth $\sigma$ and the number of samples $n=m$. As expected, RJSD approaches the true JSD by increasing the number of samples while decreasing $\sigma$. Also, increasing $\sigma$ results in a lower divergence, indicating that larger bandwidths reduce the estimator's ability to distinguish between distributions. Conversely, with a limited number of samples, as $\sigma$ approaches zero, the divergence rapidly increases and reaches its maximum value ($\log(2)$). This behavior suggests that in a learning setting, if not controlled, the divergence can be trivially maximized by decreasing $\sigma$, or equivalently, by spreading the samples across the space.

\begin{figure}
    \centering
    \includegraphics[width = 1.0\textwidth]{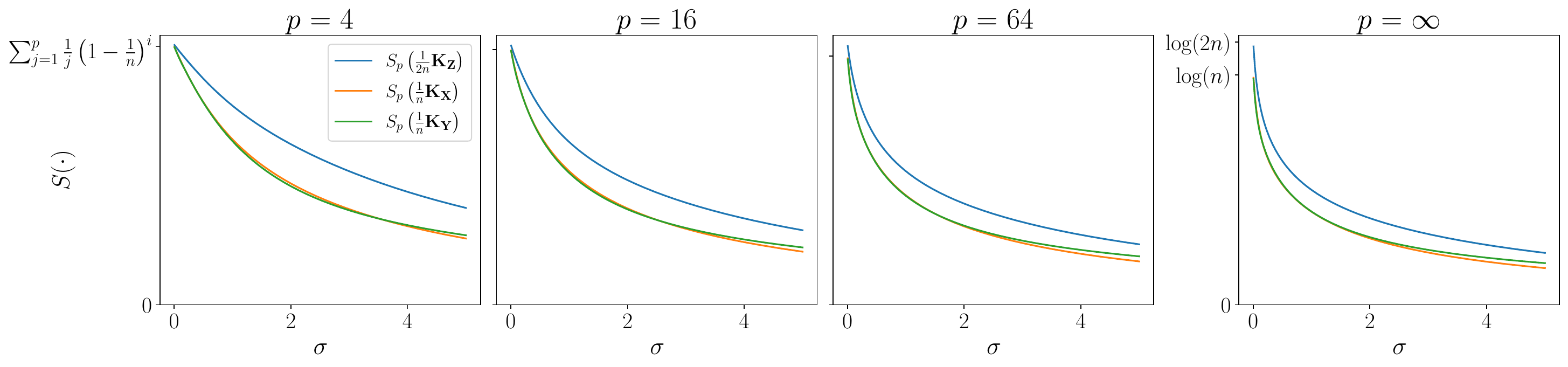}
    \caption{Approximation effect in the entropy terms with the power-series estimator of order $p$.}
    \label{fig:approximation}
\end{figure}

The proposed regularized estimators effectively prevent trivial maximization when $\sigma$ is close to zero. Fig. \ref{fig:power_series} illustrates the behavior of the power series estimator at different orders of approximation $p$. We observe that as the bandwidth increases, the divergence peaks and diminishes as $\sigma$ grows to infinity from the lowest bandwidth. Figure \ref{fig:approximation} demonstrates the regularizing effect of the approximation. We compare the three entropy terms involved in the divergence computation across different approximation orders $p$. For smaller $p$, the three entropy terms converge to similar values in the limit when $\sigma = 0$, reducing the artificial gap produced by the original entropy formula in Eqn. \ref{eq:matrix_based_entropy} ($p=\infty$) and avoiding trivial maximization of the divergence with respect to $\sigma$.

For the Fourier features-based estimator, illustrated in Fig. \ref{fig:fourier_features}, when the number of Fourier features is smaller than the number of samples, the estimator exhibits regularized behavior similar to the power-series estimator. However, when the number of features is greater or equal to the number of samples, its behavior aligns more closely with the unregularized kernel-based estimator. This behavior is expected since more Fourier features closely approximate the kernel-based estimator.

Additionally, we conduct experiments to analyze the behavior of the RJSD estimators in high-dimensional settings. For this experiment, $P \sim \mathcal{N}(\vmu_\vx, \mI_d)$ and $Q \sim \mathcal{N}(\vmu_\vy, \mI_d)$ represent two $d$-dimensional Gaussian distributions with identity covariance matrices and different means. We fix $\lVert \vmu_\vx - \vmu_\vy \rVert_2 = c$ while increasing the dimensionality of the data. 

Fig. \ref{fig:RJSD_dimensionality} shows that the original kernel-based estimator of RJSD saturates at $\log(2)$ for small $\sigma$ values, particularly rapidly in high-dimensional spaces. This behavior is undesirable as it fails to penalize sparse kernel matrices whose pairwise similarities are insufficient to accurately determine the distribution discrepancy and make the divergence susceptible to trivial maximization. In contrast, Figs. \ref{fig:power_series_dimension} and  \ref{fig:fourier_features_dimension} demonstrate that the alternative RJSD estimators do not exhibit trivial saturation at small kernel bandwidths. This property is advantageous as it effectively penalizes sparse kernel matrices, ensuring accurate measurement of the distributions' divergence even in high-dimensional settings.

% \begin{figure}
%     \centering
%     \subfigure[]{
%     \includegraphics[width=0.31\textwidth]{Figures/RJSD_dimensionality.pdf}
%     \label{fig:RJSD_dimensionality}}
%     \hspace{-0.5cm}
%     \subfigure[]{
%     \includegraphics[width=0.31\textwidth]{Figures/RJSD_taylor_dimensionality.pdf}}
%     \hspace{-0.3cm}
%     \subfigure[]{
%     \includegraphics[width=0.31\textwidth]{Figures/RJSD_dimensionality_32.pdf}}
%     % \includegraphics[width=1\linewidth]{Figures/comparison_dimensionality.pdf}
%     \caption{RJSD estimators comparison at different dimensionalities for two multivariate Gaussians with a fixed distance between their means, $n = 256$.}
%     \label{fig:RJSD estimators_dimensionality}
% \end{figure}

\subsection{Variational Estimation of Jensen-Shannon Divergence}

\begin{figure*}[!t]
    \centering
\includegraphics[width=0.9\linewidth]{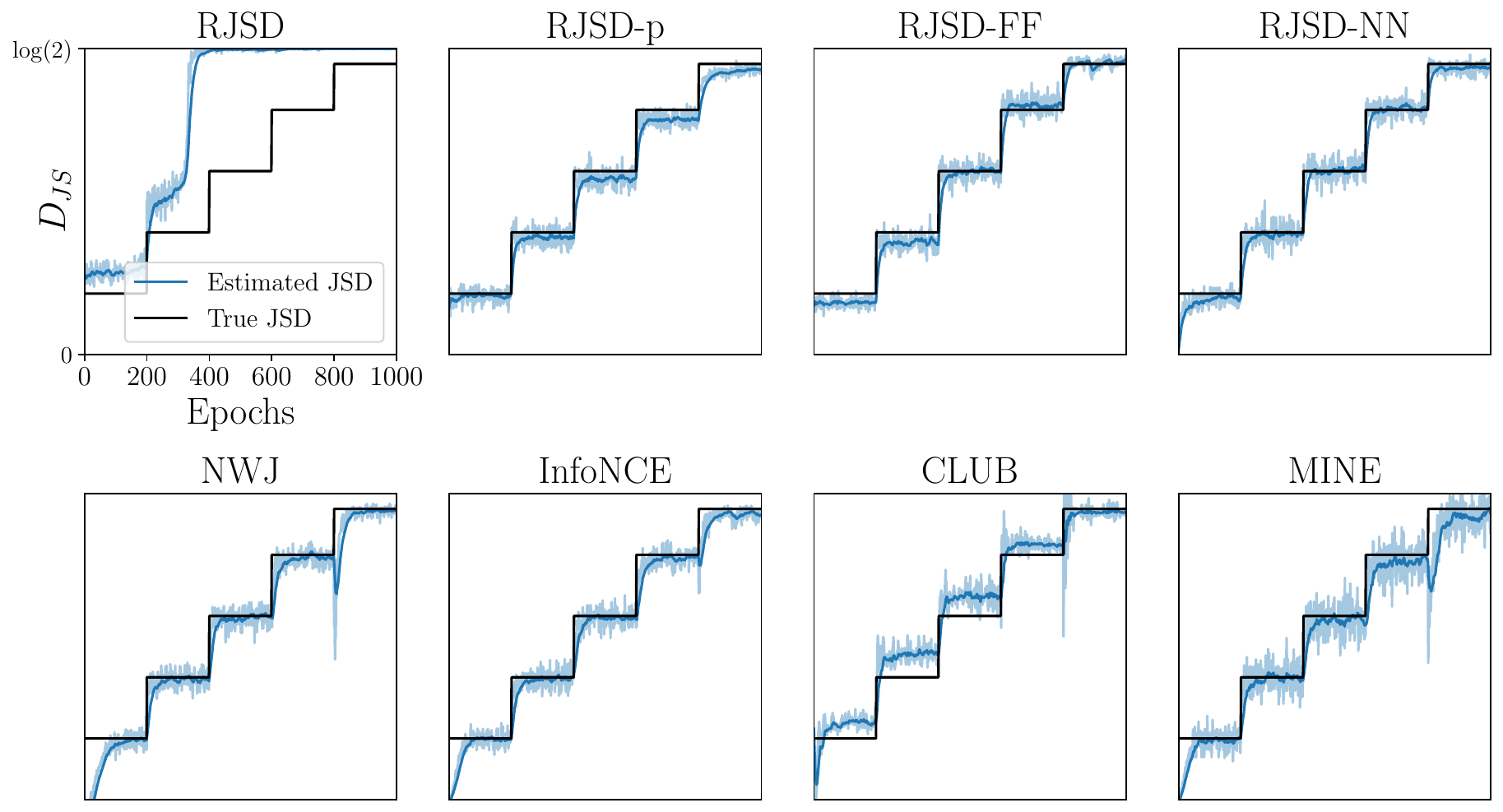}\label{fig:JSD_estimation}

\caption{Jensen-Shannon Divergence estimation for two sets of samples following Cauchy distributions (N = 512). We compare the following estimators:  kernel-based RJSD, power-series RJSD-p, Fourier Features-based RJSD-FF, Neural Network-based RJSD-NN, NWJ \citep{nguyen2010estimating}, infoNCE \citep{oord2018representation}, CLUB \citep{cheng2020club}, MINE \citep{belghazi2018mutual}. The black line is the closed-form JS divergence between the Cauchy distributions. The parameters of the distributions are changed every 200 epochs to increase the divergence.} \label{fig:JSDestimation}
\end{figure*}

\label{sec:variational_estimation}
We exploit the lower bound in Theorem \ref{th:lower_bound} to derive a variational method for estimating the classical Jensen-Shannon divergence (JSD) given only samples from $P$ and $Q$. The goal is to optimize the kernel hyper-parameters that maximize the lower bound in Eqn. \ref{eq:lower_bound}. For the kernel-based estimators, this is equivalent to finding the optimal bandwidth $\sigma$ or the $d\times d$ bandwidth matrix for a Gaussian kernel. For the Fourier Features-based estimator, we aim to optimize the Fourier features to maximize the lower bound in Eqn. \ref{th:lower_bound}. We can also optimize a neural network $f_\theta(\cdot)$ to learn a deep representation and compute the divergence of the data embedding. This formulation leads to a variational estimator of classical JSD.

\begin{definition}
(Jensen-Shannon divergence variational estimator). Let $\Theta$ be the set of all kernel hyer-parameters, and neural network weights (if utilized). We define our JSD variational estimator as: 
\begin{equation*}
    \widehat{D_{\scriptscriptstyle JS}}(P, Q) = \sup_{\theta \in \Theta}\hat{D}_{\scriptscriptstyle JS}^{\mathcal{H}_\theta}\left(f_{\theta}(X),f_\theta(Y)\right)
\end{equation*}
\end{definition}

This approach leverages the expressive power of deep networks and combines it with the capacity of kernels to embed distributions in an RKHS. This formulation allows us to model distributions with complex structures and to improve the estimator's convergence by the universal approximation properties of the neural networks \citep{wilson2016deep, liu2020learning}.

We evaluate the performance of our variational estimator of Jensen-Shannon divergence (JSD) in a tractable synthetic experiment. Here, $P(x; l_p, s_p)$ and $Q(x; l_q, s_q)$ represent two Cauchy distributions with location parameters $l_p$ and $l_q$, and scale parameters $s_p = s_q = 1$. We set $l
_p=0$ and vary the location parameter $l_q$ over time to control the target divergence. Then, we apply our variational estimator to compute JSD drawing $n = 512$ samples from both distributions at every epoch. We compare the estimates of divergence against different neural estimators. JSD corresponds to the mutual information between the mixture distribution and a Bernoulli distribution, indicating when a sample is drawn from $P$ or $Q$. Therefore, we use mutual information estimators to approach the JSD estimation, such as NWJ \citep{nguyen2010estimating}, infoNCE \citep{oord2018representation}, CLUB \citep{cheng2020club}, and MINE \citep{belghazi2018mutual}.  

Fig. \ref{fig:JSDestimation} presents the estimation results. As we expected, the original kernel-based RJSD estimator is unsuitable for this task because this estimator can be trivially maximized by decreasing the bandwidth to zero saturating at $\log(2)$. Contrarily, RJSD-p ($p=10$) succeeds in tuning the kernel bandwidth that maximizes the divergence approximating the underlying Jensen-Shannon divergence (JSD). The Fourier features-based estimator, RJSD-FF ($D=64$), optimizes the Fourier features to maximize the divergence. Alternatively, RJSD-NN optimizes a neural network $f_\theta(\cdot)$ (1 hidden layer with 64 neurons and \texttt{tanh} activation function) to learn a data representation, computing the divergence of the network output using Fourier features ($D=64$). While all compared methods approximate JSD, some exhibit high variance (MINE), bias (CLUB), or struggle to adapt to distribution shifts (InfoNCE and NWJ). Such abrupt adjustments could lead to instabilities during training. In contrast, the proposed RJSD estimators accurately estimate the divergence with lower variance, adapting seamlessly to distribution changes. 

These results highlight that RJSD is a divergence measurement that can effectively capture the underlying JSD of the original distributions and that we can learn data representations that capture the discrepancy between the original distributions by maximizing RJSD between the outputs of a deep neural network. 

% Additionally, by using Exponential Moving averages (EMA) of the covariance matrices, the estimation variance decreases further yielding a smoother estimation. Finally, we compute RJSD for a fixed set of Fourier features without any optimization (no gradients backpropagated), and we can observe that RJSD still approximates the true divergence. This result agrees with theorem \ref{th:divergence_equality} suggesting that the computed kernel implicitly approximates the underlying distributions of the data.  

\subsection{Two-sample Testing}

We evaluate the discriminatory power of RJSD for two-sample testing. Given two sets of samples, $\mX = \left\{\vx_i\right\}_{i=1}^n$  and $\mY = \left\{\vy_i\right\}_{i=1}^m$, drawn from $P$ and $Q$ respectively, two-sample testing aims to determine whether $P$ and $Q$ are identical. The null hypothesis $H_0$ states $P=Q$, while the alternative hypothesis $H_1$ states $P \neq Q$. A hypothesis test is then performed, rejecting the null hypothesis if $\mathbb{D}(P,Q)>\varepsilon$ for some distance or divergence $\mathbb{D}$ and threshold $\varepsilon>0$.  

Let $\mZ = \left\{\vz_i\right\}_{i=1}^{n+m} = \left\{\vx_1, \dots \vx_n, \vy_1, \dots, \vy_m \right\} $ be the combined sample. One common approach to perform two-sample testing is through permutation tests. These tests apply permutations of the combined data $\mZ$ to approximate the distribution of the divergence measurement under the null hypothesis. Finally, this distribution determines the rejection threshold $\varepsilon$ according to some specified significance level.

Among the most widely used metrics for two-sample testing is the maximum mean discrepancy (MMD) \citep{gretton2012kernel}. Several MMD-based tests have been proposed over the past decade \citep{gretton2012kernel,sutherland2016generative,jitkrittum2016interpretable,liu2020learning,schrab2023mmd, biggs2024mmd}. In this experiment, we employ RJSD as the divergence measure to perform hypothesis testing.

Taking inspiration from 3 well-known MMD-based tests, we designed RJSD-based versions of MMD-Split \citep{sutherland2016generative}, MMD-Deep \citep{liu2020learning}, and MMD-Fuse \citep{biggs2024mmd}. \textbf{RJSD-Split} involves splitting the data into training and testing sets to identify the optimal kernel bandwidth on the training set and subsequently evaluate performance on the testing set. Leveraging the lower bound in Eqn. \ref{eq:upper_bound}, we propose selecting the kernel hyper-parameters that maximize RJSD as these parameters enhance the distinguishability between the two distributions \citep{sutherland2016generative}. Since the kernel-based estimator is not suitable for maximization with respect to the kernel hyperparameters, we use the power-series RJSD estimator. 

Similarly, \textbf{RJSD-Deep} involves learning the parameters of the following kernel $\kappa_\theta(x,y)$:

\begin{align*}\label{eq:deep_kernel}
    \kappa_\theta(x,y) &= \left[(1-\epsilon)\kappa_1(f_\theta(x),f_\theta (y)) + \epsilon\right]\kappa_2(x,y), 
\end{align*}

where $f_\theta: \mathcal{X} \rightarrow \mathcal{F}$ represents a deep network that extracts features from the data, thereby enhancing the kernel's flexibility and its ability to capture the structure of complex distributions accurately. Here, $0 < \epsilon < 1$, and $\kappa_1$ and $\kappa_2$ are Gaussian kernels. Ultimately, we learn the network weights, the kernel bandwidths for $\kappa_1$ and $\kappa_2$, and the value of $\epsilon$ that maximizes RJSD.

On the other hand, \textbf{RJSD-Fuse} consists in combining the RJSD estimates of different kernels $\kappa \in \mathcal{K}$ drawn from a distribution $\rho \in \mathcal{M}_{+}^1(\mathcal{K})$. Then, these different values are passed through a weighted smooth maximum function that considers information from each kernel simultaneously, resulting in a new statistic. The fused statistic with parameter $\lambda > 0$ is defined as:

% \begin{equation}
% \widehat{\text{FUSE}}_{\scriptscriptstyle JS}(\mX, \mY) = \frac{1}{\lambda} \log \left(\mathbb{E}_{\kappa \sim \rho }\left[ \exp \left(\lambda \frac{\hat{D}_{\scriptscriptstyle pJS}^{\kappa}(\mX, \mY)}{\sqrt{N_\kappa(\mZ)}}\right)\right]\right),
% \end{equation}
\begin{equation*}
\widehat{\text{FUSE}}_{\scriptscriptstyle JS}(\mX, \mY) = \frac{1}{\lambda} \log \left(\mathbb{E}_{\kappa \sim \rho }\left[ \exp \left(\lambda \widehat{D}_{\scriptscriptstyle pJS}^{\: \kappa}(\mX, \mY)\right)\right]\right).
\end{equation*}

% In MMD-Fuse \citep{biggs2024mmd}, the different MMD estimates are normalized by a permutation invariant factor $N_\kappa(\mZ):= \sqrt{\frac{1}{n\times(n-1)}\sum_{i\neq j} \kappa(z_i, z_j)^2}$ to account for the different scales and variances of distinct kernels before computing the ``maximum". To include this term within our approach, instead of normalizing the divergence estimates, we normalize the kernels by $N_\kappa(\mZ)$, which in the case of $p=1$ is equivalent to MMD-Fuse (See Appendix \ref{ap:RJSD_Fuse}). 

This method does not require data-splitting since the optimal kernel is chosen unsupervised through the log-sum-exponential function. See Appendix \ref{ap:RJSD_Fuse} for implementation details.

\subsubsection{Setup}
\begin{figure}
    \centering
    \includegraphics[width=1.0\textwidth]{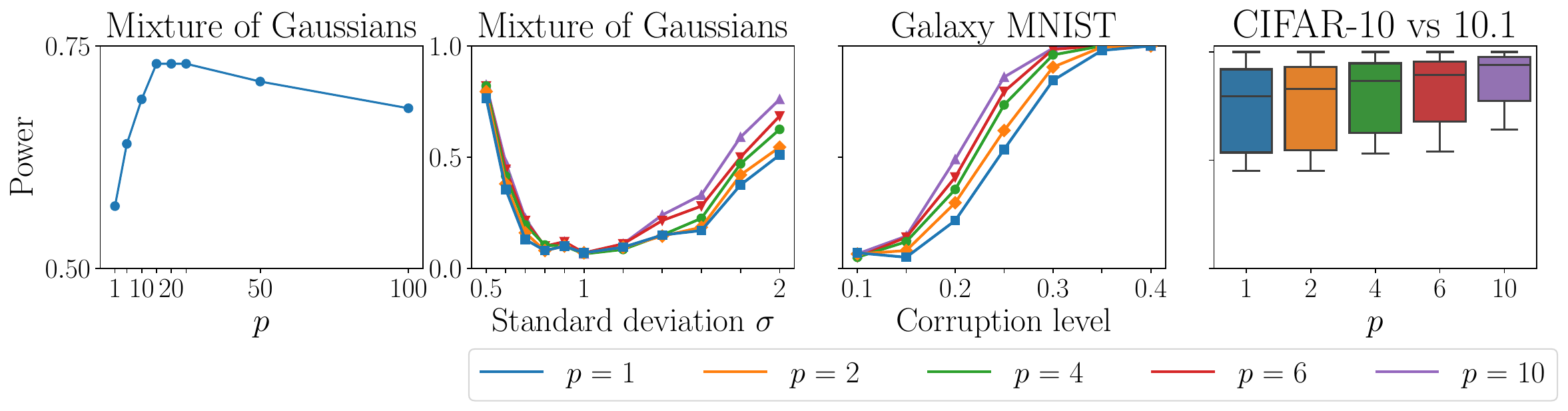}
        \caption{Test Power comparison for different orders of approximation. For the mixture of Gaussians and Galaxy MNIST, we deviate from the null hypothesis for a fixed number of samples of $n=m=500$. For CIFAR-10 vs 10.1, we show the boxplot of the distribution of the average test power for different training sets. }
    \label{fig:comparing_orders}
\end{figure}
We evaluate RJSD discriminatory power using one synthetic dataset and two real-world benchmark datasets for two-sample testing. The \textbf{Mixture of Gaussians} dataset \citep{biggs2024mmd} consists of 2-dimensional mixtures of four Gaussians $P$ and $Q$ with means at ($\pm\mu, \pm\mu$) and diagonal covariances. All components of $P$ have unit variance, while only three components of $Q$ have unit variance, with the standard deviation $\sigma$ in the fourth component being varied. The null hypothesis $H_0: \ P=Q$ corresponds to the case where $\sigma = 1$. The \textbf{Galaxy MNIST} dataset \citep{walmsley2022galaxy} consists of four categories of galaxy images captured by a ground-based telescope. $P$ represents uniformly sampled images from the first three categories, while $Q$ represents samples drawn from the first three categories with probability $1-c$ and from the fourth category with probability $c \in [0,1]$. We vary the corruption level $c$, with the null hypothesis corresponding to the case where $c = 0$. Finally, the \textbf{CIFAR 10 vs 10.1} dataset \citep{liu2020learning} compares the distribution $P$ of the original CIFAR-10 dataset \citep{krizhevsky2009learning} with the distribution $Q$ of CIFAR-10.1, which was collected as an alternative test set for models trained on CIFAR-10.

We compare the test power of RJSD-Split, RJSD-Deep, and RJSD-Fuse against various MMD-based tests:  data splitting (MMD-Split)\citep{sutherland2016generative}, Smooth Characteristic Functions (SCF) \citep{jitkrittum2016interpretable}, the MMD Deep kernel (MMD-Deep) \citep{liu2020learning}, Automated Machine Learning (AutoTST) \citep{kubler2022automl}, kernel thinning to (Aggregate) Compress Then Test (CTT \& ACTT)\citep{domingo2023compress}, and MMD Aggregated (Incomplete) tests (MMDAgg \& MMDAggInc) \citep{schrab2023mmd} and MMD-FUSE \citep{biggs2024mmd}. 

\subsubsection{Results}
\begin{figure}
    \centering
    \includegraphics[width=1.0\textwidth]{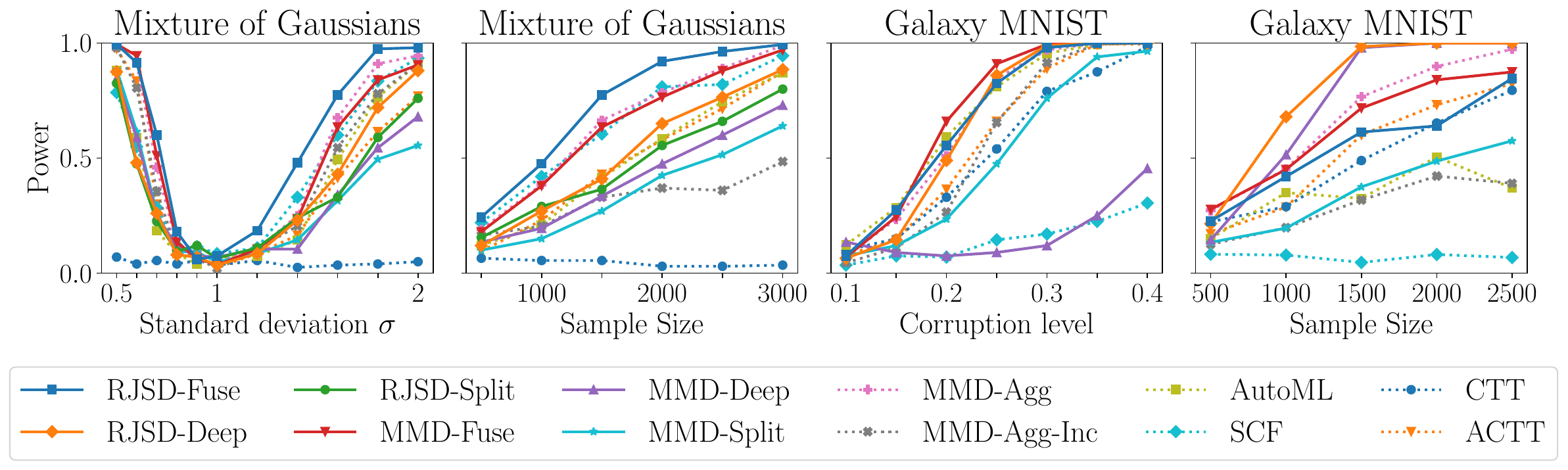}
    \caption{Test Power comparison different methods.}
    \label{fig:tst_comparison}
\end{figure}

We first investigate the impact of increasing the approximation order $p$ in the power-series expansion on test performance. Fig. \ref{fig:comparing_orders} illustrates this effect across various datasets and scenarios. For the mixture of Gaussians with a fixed standard deviation $\sigma = 2$ and $n=m=500$, we analyze the test power of RJSD-Split as $p$ increases (leftmost). The results indicate a monotonic increase in test power up to a particular order, after which it declines. This pattern was consistently observed across different standard deviations. Similarly, for the Galaxy MNIST ($n=m=500$) and CIFAR-10 vs. 10.1 ($n=m=2021$) datasets, we evaluate RJSD-Deep with varying approximation orders. The trend was consistent across all scenarios, with higher-order approximations outperforming lower ones. Notably, $p = 10$ achieved the highest test power in each case. It is important to note that $p = 1$ corresponds to MMD, highlighting that RJSD consistently exhibits superior test power compared to MMD.

Fig. \ref{fig:tst_comparison} compares the test power of various approaches across the tested datasets. In most scenarios, RJSD-Fuse ($p=10$) consistently outperforms or matches the performance of state-of-the-art methods like MMD-Fuse and MMD-Agg. Similarly, RJSD-Deep and RJSD-Split also demonstrate superior test power compared to their MMD counterparts in most cases. However, in the Galaxy MNIST dataset, when the sample size is increased, RJSD-Deep leads in performance, while RJSD-Fuse slightly falls behind MMD-Fuse. This discrepancy may be attributed to our estimator's lack of bias correction, which could affect certain cases. 

\begin{table}
\centering 

\begin{tabular}{cc}
\hline
Tests       & Power \\
\hline
\textbf{RJSD-Fuse}   & \textbf{1.000} \\
MMD-Fuse    & 0.937 \\
MMD-Agg     & 0.883 \\
\underline{RJSD-Deep}   & \underline{0.868} \\
MMD-Deep    & 0.744 \\
CTT         & 0.711 \\
ACTT        & 0.678 \\
AutoML      & 0.544 \\
MMD-Split   & 0.316 \\
MMD-Agg-Inc & 0.281 \\
SCF         & 0.171 \\
\hline
\multicolumn{2}{p{4cm}}{\raggedright \tiny \textbf{Bold:} Best approach \\
\underline{Underline:} Best data-splitting approach}  \\
\end{tabular}
\caption{Average test power for CIFAR-10 vs. CIFAR-10.1.}
\label{tab:cifar_results}
\end{table}
% \begin{wraptable}{r}{0.3\linewidth}
% \centering
% \captionof{table}{Average test power for CIFAR-10 vs. CIFAR-10.1.} 
% \label{tab:cifar_results}
% \begin{tabular}{cc}
% \hline
% Tests       & Power \\
% \hline
% \textbf{RJSD-Fuse}   & \textbf{1.000} \\
% MMD-Fuse    & 0.937 \\
% MMD-Agg     & 0.883 \\
% \underline{RJSD-Deep}   & \underline{0.868} \\
% MMD-Deep    & 0.744 \\
% CTT         & 0.711 \\
% ACTT        & 0.678 \\
% AutoML      & 0.544 \\
% MMD-Split   & 0.316 \\
% MMD-Agg-Inc & 0.281 \\
% SCF         & 0.171 \\
% \hline
% \multicolumn{2}{p{4cm}}{\raggedright \tiny \textbf{Bold:} Best approach \\
% \underline{Underline:} Best data-splitting approach}  \\
% \end{tabular}
% \end{wraptable}

Additionally, Table \ref{tab:cifar_results} presents the average power test for CIFAR-10 vs. CIFAR-10.1 computed over ten distinct training sets and 100 testing sets per training set (total of 1000 repetitions). Again, RJSD-Fuse ($p=10$) achieves the highest test power, outperforming all other methods. Also, RJSD-Deep achieves the maximum power among data-splitting techniques, significantly surpassing MMD-Deep. These results highlight the robustness and efficacy of RJSD in measuring and detecting differences in distributions, demonstrating its potential as a powerful alternative to MMD for both statistical testing and broader machine-learning applications.

\subsection{Domain Adaptation}

To test the ability of RJSD to minimize the divergence between distributions in deep learning applications, we apply RJSD to unsupervised domain adaptation. In unsupervised domain adaptation, we are given a labeled source domain $\mathcal{D}_s = \left\lbrace\vx_i^s, \vl_i^s\right\rbrace_{i=1}^{n}$ and an unlabeled target domain $\mathcal{D}_t = \left\lbrace\vx_i^t\right\rbrace_{i=1}^{m}$. The goal is to train a deep neural network $f_\theta(\cdot)$ to learn a domain-invariant representation using the source domain that allows us to infer the labels of the unsupervised target domain, that is, $f_{\theta}(\vx_i^t) \cong \vl_i^t$. 

% The source and target domain examples are sampled from joint distributions $P(\mX^s, \mL^s)$ and $Q(\mX^t, \mL^t)$.

One common approach to reducing cross-domain discrepancy is minimizing the divergence of the representation's distributions in deep layers. Let $\mathcal{L} = \left\lbrace l_1, \dots, l_{|\mathcal{L}|} \right\rbrace$ be the set of layers of a neural network where the features are not safely transferable across domains. Let $\mS^{l_1}, \dots ,\mS^{{l_{\lvert \mathcal{L}\rvert}}}$ be the source domain features and $\mT^{l_1}, \dots ,\mT^{{l_{\lvert \mathcal{L}\rvert}}}$ be the target domain features of the layers in $\mathcal{L}$. \citet{long2017deep} propose using maximum mean discrepancy (MMD) to match the joint distributions $P(\mS^{l_1}, \dots ,\mS^{{l_{\lvert \mathcal{L}\rvert}}})$ and $Q(\mT^{l_1}, \dots ,\mT^{{l_{\lvert \mathcal{L}\rvert}}})$. This methodology is known as Joint Adaptation Networks (JAN). In this experiment, we propose to use RJSD instead of MMD to explicitly minimize the divergence between the joint distributions of the activations in layers $\mathcal{L}$. Similarly to \citet{long2017deep}, we consider the joint covariance operators $\bm{C}_{\mS^{l_1: l_{\lvert\mathcal{L}\lvert}}} \in \bigotimes\limits_{l=1}^{\lvert\mathcal{L}\lvert}\mathcal{H}_l$ and $\bm{C}_{\mT^{l_1:l_{\lvert\mathcal{L}\lvert}}} \in  \bigotimes\limits_{l=1}^{\lvert\mathcal{L}\lvert}\mathcal{H}_l$, which are associated with the product of the layers' marginal kernels as follows: 

\begin{align*}
    S\left(\bm{C}_{\mS^{l_1: l_{\lvert\mathcal{L}\lvert}}}\right) = S\left(\frac{1}{n}\left[\mK_{\mS}^{l_1} \circ \mK_{\mS}^{l_2},\cdots, \circ \mK_{\mS}^{l_{\lvert\mathcal{L}\rvert}}  \right]\log \frac{1}{n}\left[\mK_{\mS}^{l_1} \circ \mK_{\mS}^{l_2}, \cdots,\circ \mK_{\mS}^{l_{\lvert\mathcal{L}\rvert}} \right] \right) \\
    S\left(\bm{C}_{\mT^{l_1:l_{\lvert\mathcal{L}\lvert}}} \right) = S\left(\frac{1}{n}\left[\mK_{\mT}^{l_1} \circ \mK_{\mT}^{l_2},\cdots, \circ \mK_{\mT}^{l_{\lvert\mathcal{L}\rvert}}  \right]\log \frac{1}{n}\left[\mK_{\mT}^{l_1} \circ \mK_{\mT}^{l_2}, \cdots,\circ \mK_{\mT}^{l_{\lvert\mathcal{L}\rvert}} \right] \right),    
\end{align*}

\noindent where $\circ$ denotes the Hadamard product. Thus, we compute the joint RJSD as: 

\begin{equation*}
    \widehat{D}_{\scriptscriptstyle JS}^\mathcal{H}(P,Q) = S\left(\frac{\bm{C}_{\mS^{l_1: l_{\lvert\mathcal{L}\lvert}}}+ \bm{C}_{\mT^{l_1:l_{\lvert\mathcal{L}\lvert}}}}{2}\right) - \frac{1}{2} \biggl(S\left(\bm{C}_{\mS^{l_1: l_{\lvert\mathcal{L}\lvert}}}\right) + S\left(\bm{C}_{\mT^{l_1:l_{\lvert\mathcal{L}\lvert}}}\right) \biggr). 
\end{equation*}

Finally, for unsupervised domain adaptation, we minimize the following loss function: 

\begin{equation*}
    \min_{\theta \in \Theta} \frac{1}{n}\sum_{i=1}^n J(f_\theta(\vx_i^s), \vl_i^s)) + \beta \widehat{D}_{\scriptscriptstyle JS}^\mathcal{H}(P,Q),
\end{equation*}

\noindent where $J(\cdot,\cdot)$ is the cross-entropy loss function, and $\beta$ is a trade-off parameter. 

Next, we evaluate the power series-based RJSD estimator ($p=3$) to minimize the cross-domain discrepancy in unsupervised domain adaptation using JANs. While other advanced domain adaptation techniques exist, our primary objective is to evaluate RJSD's effectiveness in reducing divergence between distributions in deep learning applications and compare its performance with the well-known MMD.  

\subsubsection{Setup}

We compare RJSD against MMD in 4 benchmark datasets for domain adaptation in computer vision.
\textbf{Office-31} \citep{saenko2010adapting} contains images from 31 categories and three domains: Amazon (A), Webcam (W), and DSLR (D). \textbf{Office-Home} \citep{venkateswara2017deep} consists of 65 categories across four domains: Art (Ar), Clipart (Cl), Product (Pr), and Real-World (Rw). \textbf{ImageNet-Rendition} (IN-R) \citep{hendrycks2021many} corresponds to a mix of multiple domains including art, cartoons, graffiti, origami, sculptures, and video game renditions of 200 ImageNet classes (IN-200). Finally, \textbf{ImageNet-Sketch} (IN-Sketch) \citep{wang2019learning} is a dataset of sketches of each of the 1000 ImageNet classes (IN-1k).

We implement our method based on the open-source Transfer Learning Library \textit{TLlib}\footnote{https://github.com/thuml/Transfer-Learning-Library} \citep{jiang2022transferability}. For Office-31, Office-Home, and ImageNet-R, we adapt the representations of the last two layers of a ResNet50 \citep{he2016deep}, namely the pooling and the fully connected layers. For ImageNet-Sketch, we adapt the last two layers of a ResNeXt-101. We do a grid search on a validation set to find the trade-off parameter $\beta$. For a fair comparison, all the remaining hyperparameters and configurations are kept by default according to the library implementation for both methods.      
\subsubsection{Results}
\begin{table}[t]
    \centering
    \caption{Office-31 accuracy on ResNet-50. Number of classes: 31.}
    \label{tab: results office}
    \begin{tabular}{lcccccc}
    \hline
     Method   &   A $\rightarrow$ W & D $\rightarrow$ A&  W $\rightarrow$ A & A $\rightarrow$ D & D $\rightarrow$ W & W $\rightarrow$ D\\
     \hline
     MMD  & 93.7 & 69.2 & 71.0 &\textbf{ 89.4} & \textbf{98.4} & \textbf{100.0}  \\
     RJSD  & \textbf{94.8} & \textbf{70.3} & \textbf{71.2} & 88.4 & 98.2 & 99.6  \\
     \hline
    \end{tabular}

\end{table}

\addtolength{\tabcolsep}{-4pt}
\begin{table}[t]
        \caption{Office-Home accuracy on ResNet-50. Number of classes: 65}
    \label{tab:results home}
    \centering
    \resizebox{\columnwidth}{!}{%
    \begin{tabular}{lcccccccccccc}
    \hline
     Method   &\small Ar{$\scriptscriptstyle \rightarrow$}Cl & \small Ar{$\scriptscriptstyle \rightarrow$}Pr & \small Ar{$\scriptscriptstyle \rightarrow$}Rw & \small Cl{$\scriptscriptstyle \rightarrow$}Ar& \small Cl{$\scriptscriptstyle \rightarrow$}Pr& \small Cl{$\scriptscriptstyle \rightarrow$}Rw&\small  Pr{$\scriptscriptstyle \rightarrow$}Ar & \small Pr{$\scriptscriptstyle \rightarrow$}Cl & \small Pr{$\scriptscriptstyle \rightarrow$}Rw &\small Rw{$\scriptscriptstyle \rightarrow$}Ar & \small Rw{$\scriptscriptstyle \rightarrow$}Cl & \small Rw{$\scriptscriptstyle \rightarrow$}Pr\\
     \hline
     MMD & 50.8 & 71.9& 76.5&\textbf{60.6} & 68.3&68.7 &60.5 &49.6 & 76.9&71.0 &55.9 & 80.5\\
     RJSD & \textbf{51.3} & \textbf{72.0}& \textbf{77.2} & 59.9 & \textbf{70.4} &\textbf{69.0}  &  \textbf{61.8}& \textbf{50.7}& \textbf{77.9}&\textbf{73.2} & \textbf{58.1} & \textbf{82.1}\\
     \hline
 
    \end{tabular}%
    }

\end{table}

Tables \ref{tab: results office}, \ref{tab:results home}, and \ref{tab: results imagenet} present the results for the tested datasets. RJSD generally outperforms MMD in most transfer tasks across all four datasets, demonstrating its effectiveness in joint distribution adaptation. Notably, RJSD significantly improves classification accuracy on ImageNet-R (IN-R), which is considered the most challenging dataset due to its mixture of multiple domains. Similarly, in ImageNet-Sketch, RJSD surpasses MMD, highlighting its ability to minimize distribution divergence even in high-dimensional spaces with many classes. 

The encouraging results achieved by RJSD underscore the potential of this quantity for divergence minimization tasks and position RJSD as a promising alternative to MMD in deep learning applications.

\begin{table}[!t]
    \centering
    \caption{ImageNet-R (IN-R) and ImageNet-Sketch (IN-Sketch) accuracies.  IN-R number of classes: 200, source domain: ImageNet-200, architecture: ResNet-50. IN-Sketch number of classes: 1000, source domain: ImageNet-1k, architecture: ResNext101.}
    \label{tab: results imagenet}
    \begin{tabular}{lcc}
    \hline
     Method   &  IN-200 $\rightarrow$ IN-R & IN-1k $\rightarrow$ IN-Sketch \\\hline
     MMD & 41.7 & 80.3 \\
     RJSD & \textbf{45.5} & \textbf{81.8} \\  
     \hline
    \end{tabular}

\end{table}
\section{Conclusions and Future Work}

% \paragraph{Limitations: }
% \begin{itemize}
%     \item We need to better understand the estimator's bias and variance. 
%     \item Faster approximations. 
% \end{itemize}

In this work, we have introduced the representation Jensen-Shannon divergence (RJSD), a novel divergence measure that leverages covariance operators in reproducing kernel Hilbert spaces (RKHS) to capture discrepancies between probability distributions. Unlike traditional methods that rely on Gaussian assumptions or density estimation, RJSD directly represents input distributions through uncentered covariance operators in RKHS, providing a flexible approach to divergence estimation.

We developed several estimators for RJSD that can be computed using kernel matrices and explicit covariance matrices from Fourier Features. We also proposed a variational method for estimating the classical Jensen-Shannon divergence by optimizing kernel hyperparameters or neural network representations to maximize RJSD. Through extensive experiments involving divergence maximization and minimization,  RJSD demonstrated superiority over state-of-the-art methods in tasks such as two-sample testing, distribution shift detection, and unsupervised domain adaptation.

The empirical results indicate that RJSD displays higher discriminative power in two-sample testing scenarios than similar MMD-based approaches. These results position RJSD as a robust alternative to traditional methods like MMD, which are widely used in the machine learning community. RJSD's versatility and effectiveness underscore its potential to become a foundational tool in machine learning research and applications.

Future work will focus on further exploring the bias and variance of the RJSD estimator and developing faster approximations to enhance its computational efficiency. Additionally, investigating RJSD’s application in broader machine learning domains could provide further insights into its utility and versatility.

\section*{Acknowledgments}

This material is based upon work supported by the Office of the Under Secretary of Defense for Research and Engineering under award number FA9550-21-1-0227.

% \url{https://openreview.net/forum?id=7GCRhebJEr}

% Domain adaptation: \url{https://openaccess.thecvf.com/content_cvpr_2018/papers/Zhang_Aligning_Infinite-Dimensional_Covariance_CVPR_2018_paper.pdf}

% "Although (Gretton et al., 2012) shows that an universal RKHS is rich enough, such kernel-based MMD may suffer from vanishing gradients for low-bandwidth kernels. Moreover, it may be possible that some widely-used kernels are unable to capture
% very complex distances in high dimensional spaces such as natural images (Reddi et al., 2015; Arjovsky et al., 2017)." \url{https://proceedings.mlr.press/v70/long17a/long17a.pdf#page=2.80}

\vskip 0.2in
\bibliography{JMLR/refs}

\newpage

\appendix
\section{}
\label{app:theorem}

\subsection{Proof Lemma \ref{th:upper_bound}} \label{ap:proof_upper_bound}
\begin{proof}
    
To prove this Lemma, we use (Proposition 4.e) in \citet{bach2022information}. We have that
\begin{equation*}
    D_{\scriptscriptstyle KL}\left(\CP , \CQ \right) \geq \frac{1}{2}\Vert \CP - \CQ \Vert_{*}^2 \geq \frac{1}{2}\Vert  \CP - \CQ \Vert_{\textrm{HS}}^2,
\end{equation*}
where $D_{\scriptscriptstyle KL}^{\scriptscriptstyle \mathcal{H}}$ is the kernel Kullback-Leibler divergence and $\lVert \cdot \rVert_*$ and $\lVert \cdot \rVert_{HS}$ denote the nuclear and Hilbert-Schmidt norms respectively. Let $\CM = \frac{\CP + \CQ}{2}$, then:
\begin{align*}
    D_{\scriptscriptstyle JS}(\CP,\CQ)  = &  \frac{1}{2}D_{\scriptscriptstyle KL}^{\scriptscriptstyle \mathcal{H}}(\CP,\CM) + \frac{1}{2}D_{\scriptscriptstyle KL}^{\scriptscriptstyle \mathcal{H}}(\CP,\CM) \\
     \geq & \frac{1}{4}\left\Vert \CP - \frac{1}{2}(\CP + \CQ) \right\Vert_{*}^2 + \frac{1}{4}\left\Vert \CQ - \frac{1}{2}(\CP + \CQ) \right\Vert_{*}^2 \\
     \geq & \frac{1}{4}\left\Vert \frac{1}{2}\CP - \frac{1}{2}\CQ \right\Vert_{*}^2 + \frac{1}{4}\left\Vert \frac{1}{2}\CQ - \frac{1}{2}\CP \right\Vert_{*}^2 = \frac{1}{8}\left\Vert \CP - \CQ \right\Vert_{*}^2
\end{align*}
 and thus, $D_{\scriptscriptstyle JS}^{\scriptscriptstyle \mathcal{H}}(\CP,\CQ) \geq  \frac{1}{8}\left\Vert \CP - \CQ \right\Vert_{*}^2 \geq \frac{1}{8}\left\Vert \CP - \CQ \right\Vert_{\textrm{HS}}^2$.

Now, let $\phi : \mathcal{X} \mapsto \mathcal{H}$ then, and $\left\{ e_{\alpha} \right\}$ be an orthonormal basis in $\mathcal{H}$, we have that
\begin{align*}
    \Tr \left(\phi(x)\otimes \phi(x) \phi(y)\otimes \phi(y)\right) = & \sum\limits_{\alpha}\langle \phi(x)\otimes \phi(x) \phi(y)\otimes \phi(y) e_{\alpha}, e_{\alpha} \rangle \\
    = & \sum\limits_{\alpha}\langle \phi(x)\langle \phi(x), \phi(y)\otimes \phi(y) e_{\alpha}\rangle , e_{\alpha} \rangle\\
    = & \sum\limits_{\alpha}\langle \phi(x)\langle \phi(x), \phi(y) \langle \phi(y), e_{\alpha}\rangle \rangle , e_{\alpha} \rangle\\
    = & \sum\limits_{\alpha}\langle \phi(x)\langle \phi(x), \phi(y)\rangle \langle \phi(y), e_{\alpha}\rangle, e_{\alpha} \rangle\\
    = & \sum\limits_{\alpha}\langle \phi(x), e_{\alpha}\rangle \langle \phi(x), \phi(y)\rangle \langle \phi(y), e_{\alpha}\rangle\\
    = & \langle \phi(x), \phi(y)\rangle \sum\limits_{\alpha}\langle \phi(x), e_{\alpha}\rangle  \langle \phi(y), e_{\alpha}\rangle = \langle \phi(x), \phi(y)\rangle \langle \phi(x), \phi(y)\rangle\\
    = & \langle \phi(x), \phi(y)\rangle^2 = \kappa(x,y)^2
\end{align*}
Note that for $T : \mathcal{H} \mapsto \mathcal{H}$, $\Tr(T^* T) = \sum_{\alpha}\langle T e_{\alpha}, Te_{\alpha}\rangle = \Vert T \Vert_{\textrm{HS}}^2$. In particular, if we have that $T = \phi(x)\otimes \phi(x) - \phi(y)\otimes \phi(y)$,
\begin{align*}
    \Vert \phi(x)\otimes \phi(x) - \phi(y)\otimes \phi(y) \Vert_{\textrm{HS}}^2 = & \Tr(\phi(x)\otimes \phi(x)\phi(x)\otimes \phi(x)) - 2\Tr(\phi(x)\otimes \phi(x)\phi(y)\otimes \phi(y))\\ 
    & + \Tr(\phi(y)\otimes \phi(y)\phi(y)\otimes \phi(y))\\
    = & \kappa^2(x,x) - 2\kappa^2(x,y) + \kappa^2(y,y)
\end{align*}
Finally, note that \begin{align*}
    \Vert \CP - \CQ \Vert_{\textrm{HS}}^2 = & \Tr(\E_{P }[\phi(x)\otimes \phi(x)]\E_{P'}[\phi(x)\otimes \phi(x)]) - 2\Tr(\E_{P}[\phi(x)\otimes \phi(x)]\E_{Q}[\phi(y)\otimes \phi(y)])\\ 
    & + \Tr(\E_{Q}[\phi(y)\otimes \phi(y)]\E_{Q'}[\phi(y)\otimes \phi(y)])\\
    = & \Tr(\E_{P,P'}[\phi(x)\otimes \phi(x)\phi(x')\otimes \phi(x')]) - 2\Tr(\E_{P, Q}[\phi(x)\otimes \phi(x)\phi(y)\otimes \phi(y)])\\ 
    & + \Tr(\E_{Q, Q'}[\phi(y)\otimes \phi(y)\phi(y')\otimes \phi(y')])\\
    = & \E_{P,P'}[\kappa^2(x,x')] - 2\E_{P,Q}[\kappa^2(x,y)] + \E_{Q,Q'}[\kappa^2(y,y')],
\end{align*}
which corresponds to squared $\operatorname{MMD}$ with kernel $\kappa^2(\cdot, \cdot)$.
\end{proof}

\subsection{Proof Theorem \ref{th:lower_bound}}\label{ap:proof_lower_bound}
\begin{proof} For \eqref{eq:lower_bound} we have the following
\begin{align*}
    D_{\scriptscriptstyle JS}(\CP,\CQ) = & \frac{1}{2}D_{\scriptscriptstyle KL}^{\scriptscriptstyle \mathcal{H}}(\CP,\CM) + \frac{1}{2}D_{\scriptscriptstyle KL}^{\scriptscriptstyle \mathcal{H}}(\CQ,\CM) \\
     = & \frac{1}{2}D_{\scriptscriptstyle KL}^{\scriptscriptstyle \mathcal{H}}\left(\int_{\mathcal{X}}\phi(x)\otimes \phi(x) \d P(x) , \int_{\mathcal{X}}\phi(x)\otimes \phi(x) \d M(x)\right) + \\
     & + \frac{1}{2}D_{\scriptscriptstyle KL}^{\scriptscriptstyle \mathcal{H}}\left(\int_{\mathcal{X}}\phi(x)\otimes \phi(x) \d Q(x) , \int_{\mathcal{X}}\phi(x)\otimes \phi(x) \d M(x)\right)\\
     = & \frac{1}{2}D_{\scriptscriptstyle KL}^{\scriptscriptstyle \mathcal{H}}\left(\int_{\mathcal{X}}\phi(x)\otimes \phi(x) \d P(x) , \int_{\mathcal{X}}\frac{\d M}{\d P}(x)\phi(x)\otimes \phi(x) \d P(x)\right) + \\
     & + \frac{1}{2}D_{\scriptscriptstyle KL}^{\scriptscriptstyle \mathcal{H}}\left(\int_{\mathcal{X}}\phi(x)\otimes \phi(x) \d Q(x) , \int_{\mathcal{X}}\frac{\d M}{\d Q}(x)\phi(x)\otimes \phi(x) \d P(x)\right).\\ 
\end{align*}
Since $D_{\scriptscriptstyle KL}^{\scriptscriptstyle \mathcal{H}}$ is jointly convex \citep{bach2022information}, then 
\begin{align*}
    D_{\scriptscriptstyle JS}(\CP,\CQ) \leq & \frac{1}{2}\int_{\mathcal{X}} D_{\scriptscriptstyle KL}\left(\phi(x)\otimes \phi(x)  ,\frac{\d M}{\d P}(x)\phi(x)\otimes \phi(x) \right) \d P(x) + \\ 
    & + \frac{1}{2}\int_{\mathcal{X}} D_{\scriptscriptstyle KL}\left(\phi(x)\otimes \phi(x)  ,\frac{\d M}{\d Q}(x)\phi(x)\otimes \phi(x) \right) \d Q(x). \\ 
\end{align*}
Notice that $\phi(x)\otimes \phi(x)$ is a rank-1 covariance operator with one eigenvalue equal $\|\phi(x)\|^2 = 1 $ and one eigen vector $\phi(x)$, therefore, it can be simplified as:
\begin{align*}
    D_{\scriptscriptstyle JS}(\CP,\CQ) \leq & \frac{1}{2}\int_{\mathcal{X}} D_{\scriptscriptstyle KL}\left(1  , \frac{\d M}{\d P}(x)\right) \d P(x) + \frac{1}{2}\int_{\mathcal{X}} D_{\scriptscriptstyle KL}\left(1  , \frac{\d M}{\d Q}(x)\right) \d Q(x) \\ 
    = & \frac{1}{2}\int_{\mathcal{X}} D_{\scriptscriptstyle KL}\left(1 ,  \frac{\d M}{\d P}(x)\right) \d P(x) + \frac{1}{2}\int_{\mathcal{X}} D_{\scriptscriptstyle KL}\left(1  , \frac{\d M}{\d Q}(x)\right) \d Q(x) \\ 
    = & \frac{1}{2}\int_{\mathcal{X}} -\log\left(\frac{\d M}{\d P}(x)\right) \d P(x) + \frac{1}{2}\int_{\mathcal{X}} -\log\left(\frac{\d M}{\d Q}(x)\right) \d Q(x) \\ 
    & =  \frac{1}{2}D_{\scriptscriptstyle KL}(P,M) + \frac{1}{2}D_{\scriptscriptstyle KL}(Q,M)
    = D_{\scriptscriptstyle JS}(P,Q)
\end{align*}
\end{proof}

\subsection{Convergence of RJSD kernel-based estimator}
\label{ap:proof_convergence_kernel}

Since RJSD corresponds to the empirical estimation of three different covariance operator entropies, and assuming $n=m$ for simplicity, it is straightforward to show that: 

    \begin{equation*}
        \mathbb{E}\left[\widehat{D}_{\scriptscriptstyle JS}^{\: \kappa}(\mX,\mY) - D_{\scriptscriptstyle JS}^\mathcal{H}(P, Q) \right] \leq 3\left[\frac{1+c(8\log(n))^2}{\alpha n} + \frac{17}{\sqrt{n}}(2\sqrt{c} + \log(n))\right]. 
    \end{equation*}

Therefore, we can conclude that $\widehat{D}_{\scriptscriptstyle JS}^{\: \kappa}(\mX,\mY)$ converges to the population quantity $D_{\scriptscriptstyle JS}^\mathcal{H}(P, Q)$ at a rate $\mathcal{O}\left(\frac{1}{\sqrt{n}}\right)$.

\section{Two-sample testing implementation details}
\label{ap:RJSD_tst}
% \begin{figure}[!t]
%     \centering
% \includegraphics[width=0.6\textwidth]{Figures/scatter_mixture.pdf}

% \caption{Mixture of Gaussians with standard deviation $\sigma = 2$ in the the lower left mode.}
% \label{fig:mixture_samples}
% \end{figure}
\begin{figure}[!t]
\centering
\subfigure[Mixture of Gaussians]{
\includegraphics[width=0.48 \textwidth]{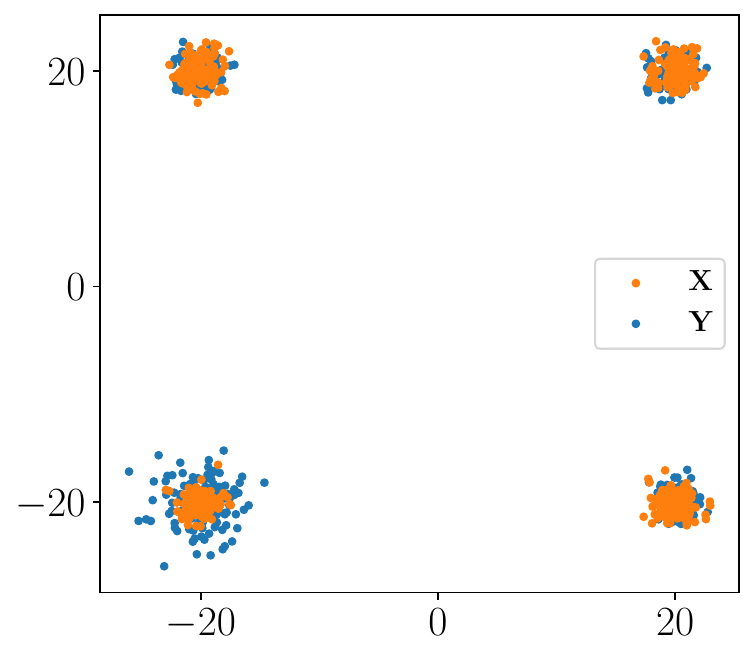}}
% \hspace{-0.3cm}
\hfill
\subfigure[Galaxy MNIST]{
\includegraphics[width=0.43\textwidth]{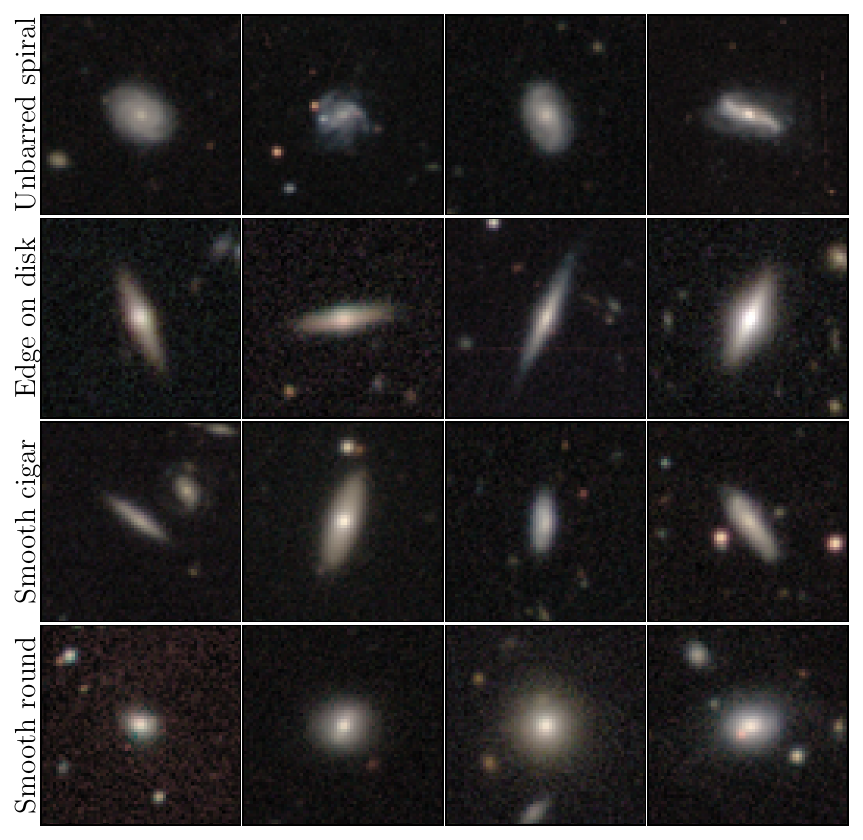}}

\caption{Mixture of Gaussians and Galaxy MNIST datasets.}
\label{fig:galaxy_Data}
\end{figure}

\begin{figure}[!t]
    \centering
    \subfigure[CIFAR 10]{\includegraphics[width=0.49\textwidth]{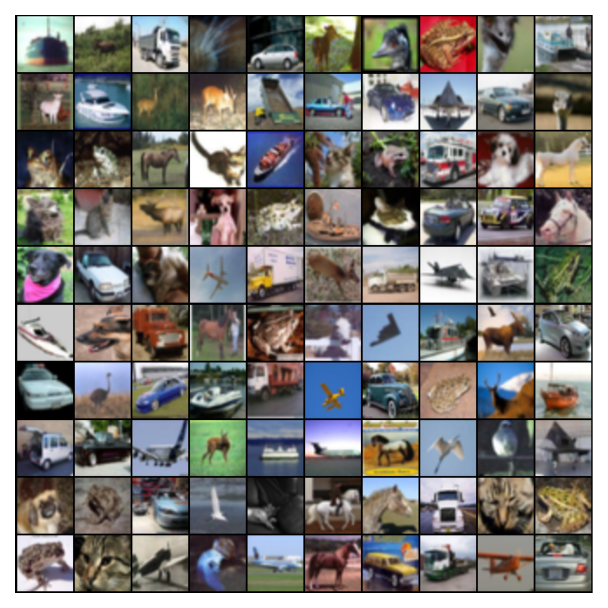}}\hspace{-0.5cm}
    \subfigure[CIFAR-10.1]{\includegraphics[width=0.49\textwidth]{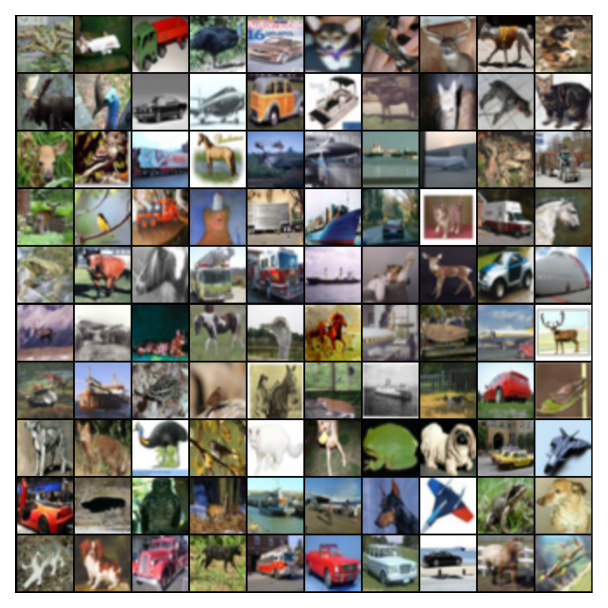}}

\caption{CIFAR 10 vs 10.1 images.}
\label{fig:CIFAR_images}
\end{figure}
Upon the paper's acceptance, all the code and model hyperparameters, including learning rates, epochs, kernel bandwidth initialization, and batch size to reproduce the results, will be uploaded. 
\subsection{RJSD-Deep}
For RJSD-Deep, we use the same model as MMD-Deep \citep{liu2020learning}, except that we removed the batch normalization layers and added a \texttt{tanh} activation function at the output of the last linear layer. 

\label{ap:RJSD_deep}
\subsection{RJSD-Fuse}
\label{ap:RJSD_Fuse}

\citet{biggs2024mmd} proposes MMD-Fuse, which computes a weighted smooth maximum of different MMD values from different kernels $\kappa \in \mathcal{K}$ drawn from a distribution $\rho \in \mathcal{M}_{+}^1(\mathcal{K})$. The proposed statistic is defined as: 

\begin{equation*}
\widehat{\text{FUSE}}_{\scriptscriptstyle MMD}(\mX, \mY) = \frac{1}{\lambda} \log \left(\mathbb{E}_{\kappa \sim \rho }\left[ \exp \left(\lambda \frac{\widehat{\operatorname{MMD}}_{\kappa}^2 (\mX, \mY)}{N_\kappa(\mZ)}\right)\right]\right).
\end{equation*}

Here, the different MMD estimates are normalized by a permutation invariant factor $N_\kappa(\mZ):= \sqrt{\frac{1}{n\times(n-1)}\sum_{i\neq j} \kappa(z_i, z_j)^2}$ to account for the different scales and variances of distinct kernels before computing the ``maximum". To include this term within our approach, instead of normalizing the divergence estimates, we normalize the kernels by $N_\kappa(\mZ)$, which in the case of $p=1$ is equivalent to MMD-Fuse. That is: 

   \begin{equation*}
    \hat{D}_{\scriptscriptstyle pJS}^{\mathcal{H}}(P,Q) = S_p\left(\tfrac{1}{n+m}\tfrac{\Kz}{\sqrt{N_\kappa(\mZ)}}\right) - \left(\tfrac{n}{n+m}S_p\left(\tfrac{1}{n}\tfrac{\Kx}{\sqrt{N_\kappa(\mZ)}}\right) + \tfrac{m}{n+m}S_p\left(\tfrac{1}{m}\tfrac{\Ky}{\sqrt{N_\kappa(\mZ)}}\right)\right).
\end{equation*} 

Notice that for $p=1$, this is equivalent to MMD-Fuse, where the measurement is normalized. However, normalizing the kernel allows the normalization to account for higher-order interactions between the kernel matrices for $p>1$. 

\paragraph{Distribution over kernels: }

\begin{figure}[!t]

\subfigure[]{
    \includegraphics[width=0.445\textwidth]{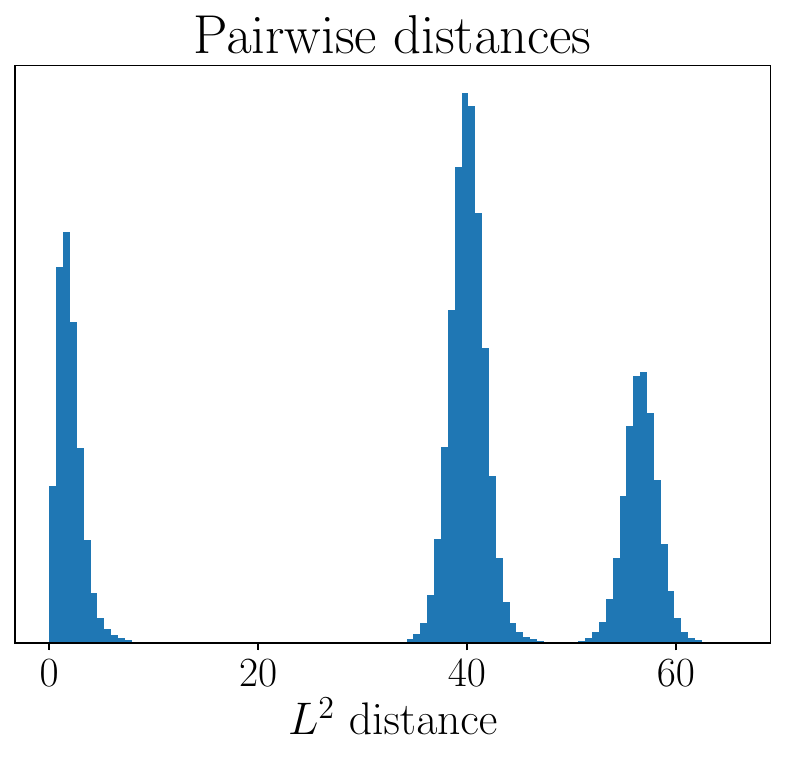}
    }
    \hspace{-0.4cm}
    \subfigure[]{
    \includegraphics[width=0.48\textwidth]{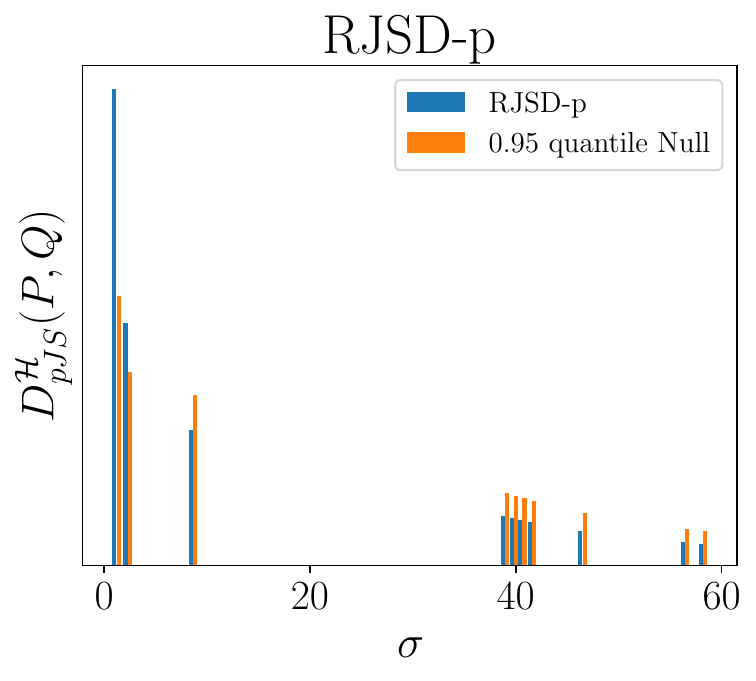}}
\caption{$\normltwo$ distance distribution for the mixture of Gaussians and RJSD estimates for ten different bandwidths tested. }
\label{fig:rjsd_fuse_mixture}
\end{figure}

Similarly to MMD-Fuse, we use a collection of Laplacian $\kappa^l_\sigma (x,x') = \exp\left(-\tfrac{\lVert x - x'\rVert_1}{\sigma}\right)$ and Gaussian $\kappa^g_\sigma (x,x') = \exp\left(-\tfrac{\lVert x - x'\rVert_2^2}{2\sigma^2}\right)$ kernels with distinct bandwidths $\sigma > 0$. In our implementation, we choose the bandwidths as the $5\%, 15\%, 25\% , \dots 95\%$ quantiles of $\left\lbrace \lVert z - z'\rVert_r: z, z' \in \mZ \right\rbrace$, with $r\in{1,2}$ for the Laplace and Gaussian kernels respectively. This choice is similar to MMD-Fuse, where ten bandwidths per kernel type are also selected. See Fig. \ref{fig:rjsd_fuse_mixture}.    

\end{document}